\documentclass[lettersize,journal]{IEEEtran}
\usepackage{amsmath,amsfonts}
\usepackage{algorithmic}
\usepackage{algorithm}
\usepackage{array}
\usepackage{textcomp}
\usepackage{stfloats}
\usepackage{url}
\usepackage{verbatim}
\usepackage{graphicx}
\usepackage{cite}
\usepackage{multirow}
\usepackage{caption}
\usepackage{subcaption}
\usepackage{bbding}
\usepackage{utfsym}
\usepackage{amsmath}
\usepackage{amsfonts}
\usepackage{array}
\usepackage{tabularx}
\usepackage{booktabs} 
\usepackage{wrapfig}
\usepackage{picinpar}
\usepackage{cutwin}
\usepackage{hyperref}
\begin{document}

\title{Unveiling the Superior Paradigm: A Comparative Study of Source-Free Domain Adaptation and Unsupervised Domain Adaptation}

\author{Fan Wang, Zhongyi Han$^{*}$, Xingbo Liu, Xin Gao, Yilong Yin$^{*}$

\thanks{This work is supported by the National Natural Science Foundation of China (U23A20389, 62176139, 62206160), the Major Basic Research Project of Natural Science Foundation of Shandong Province (ZR2021ZD15), the Young Elite Scientists Sponsorship Program by CAST, the Young Talent of Lifting Engineering for Science and Technology in Shandong (SDAST2024QTA020), the Natural Science Foundation of Shandong Province (ZR2022QF082) and special funds for distinguished professors of Shandong Jianzhu University.}

\thanks{Fan Wang and Yilong Yin are with the school of software, shandong university, Jinan, 250100, China (e-mail: fanwang@mail.sdu.edu.cn;  ylyin@sdu.edu.cn); Xingbo Liu is with the school of computer science and technology, Shandong
Jianzhu University, Jinan, 250101, China (e-mail: sclxb@mail.sdu.edu.cn); Zhongyi Han and Xin Gao are with the Computer Science Program, Computer Electrical and Mathematical Sciences and Engineering Division, Center of Excellence on Smart Health, and Center of Excellence for Generative AI, King Abdullah University of Science and Technology (KAUST), Thuwal, 23955-6900, Saudi Arabia (e-mail: zhongyi.han@kaust.edu.sa, xin.gao@kaust.edu.sa).}

\thanks{* denotes the corresponding authors}
}

\maketitle

\begin{abstract}
In domain adaptation, there are two popular paradigms: Unsupervised Domain Adaptation (UDA), which aligns distributions using source data, and Source-Free Domain Adaptation (SFDA), which leverages pre-trained source models without accessing source data. Evaluating the superiority of UDA versus SFDA is an open and timely question with significant implications for deploying adaptive algorithms in practical applications. In this study, we demonstrate through predictive coding theory and extensive experiments on multiple benchmark datasets that SFDA generally outperforms UDA in real-world scenarios. Specifically, SFDA offers advantages in time efficiency, storage requirements, targeted learning objectives, reduced risk of negative transfer, and increased robustness against overfitting. Notably, SFDA is particularly effective in mitigating negative transfer when there are substantial distribution discrepancies between source and target domains. Additionally, we introduce a novel data-model fusion scenario, where data sharing among stakeholders varies (e.g., some provide raw data while others provide only models), and reveal that traditional UDA and SFDA methods do not fully exploit their potential in this context. To address this limitation and capitalize on the strengths of SFDA, we propose a novel weight estimation method that effectively integrates available source data into multi-SFDA (MSFDA) approaches, thereby enhancing model performance within this scenario. This work provides a thorough analysis of UDA versus SFDA and advances a practical approach to model adaptation across diverse real-world environments.
\end{abstract}

\begin{figure*}[t]
\begin{minipage}[b]{.245\linewidth}
\centering
\includegraphics[height=4cm,width=4.4cm]{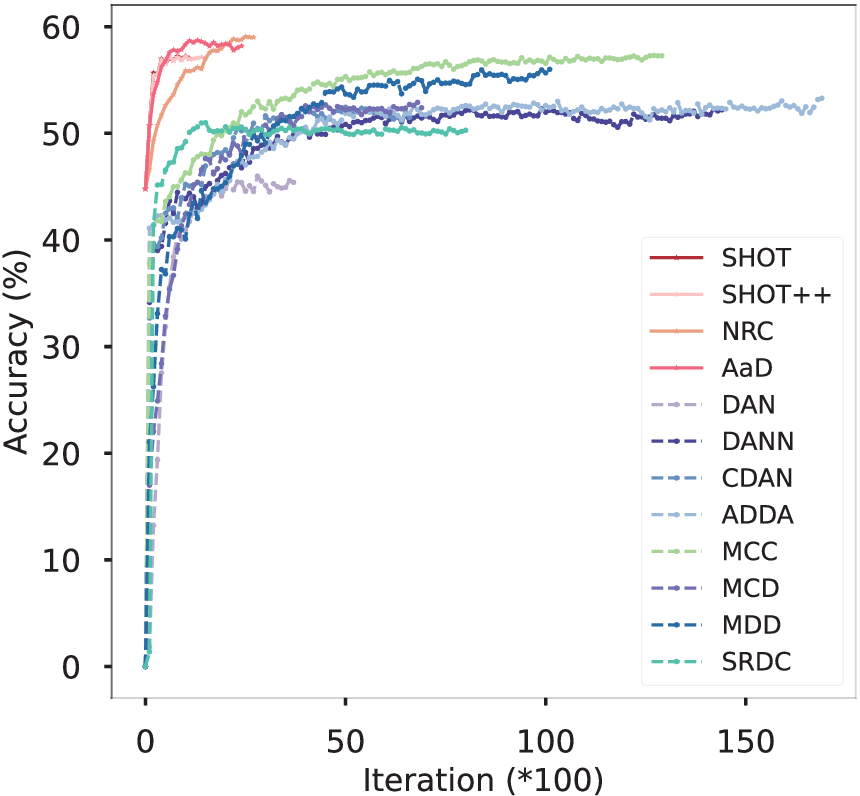}
\subcaption{Ar$\rightarrow$Cl}
\end{minipage}
\begin{minipage}[b]{.245\linewidth}
\centering
\includegraphics[height=4cm,width=4.4cm]{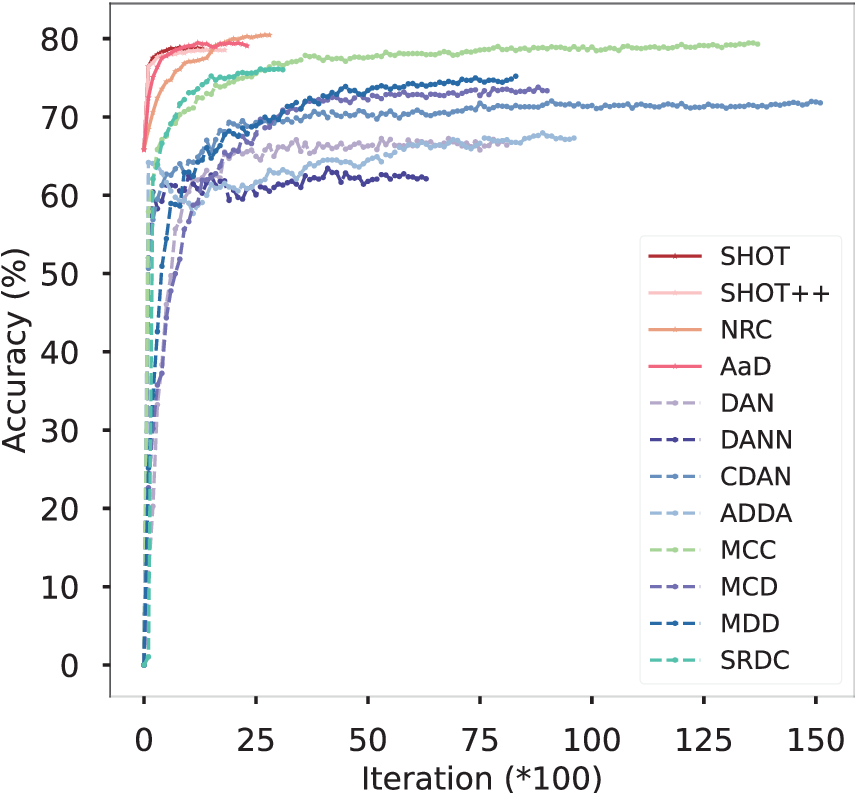}
\subcaption{Ar$\rightarrow$Pr}
\end{minipage}
\begin{minipage}[b]{.245\linewidth}
\centering
\includegraphics[height=4cm,width=4.4cm]{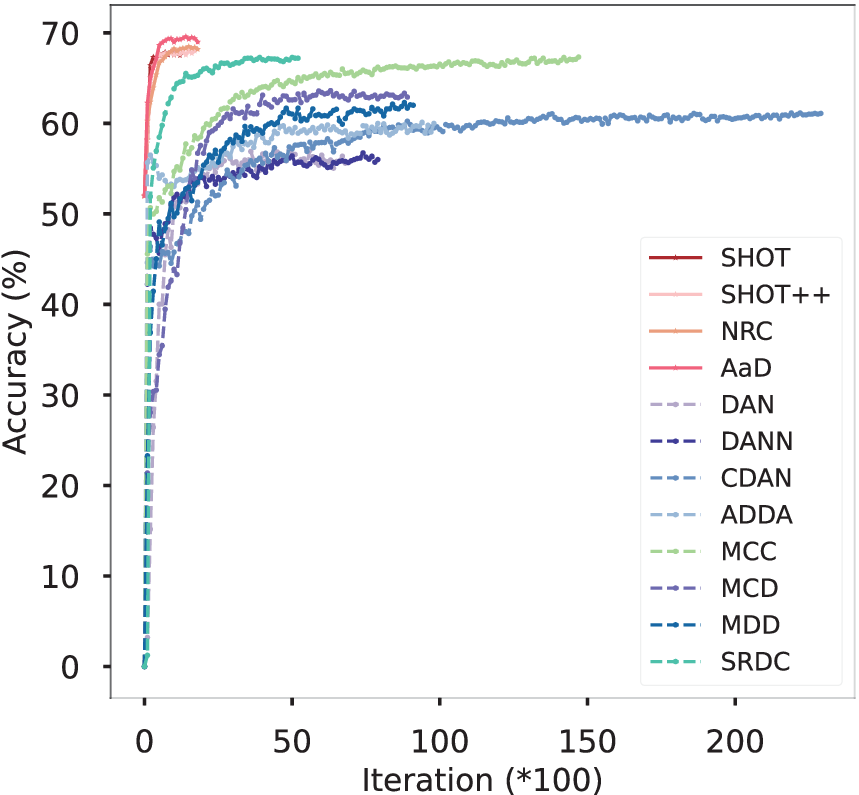}
\subcaption{Cl$\rightarrow$Ar}
\end{minipage}
\begin{minipage}[b]{.245\linewidth}
\centering
\includegraphics[height=4cm,width=4.4cm]{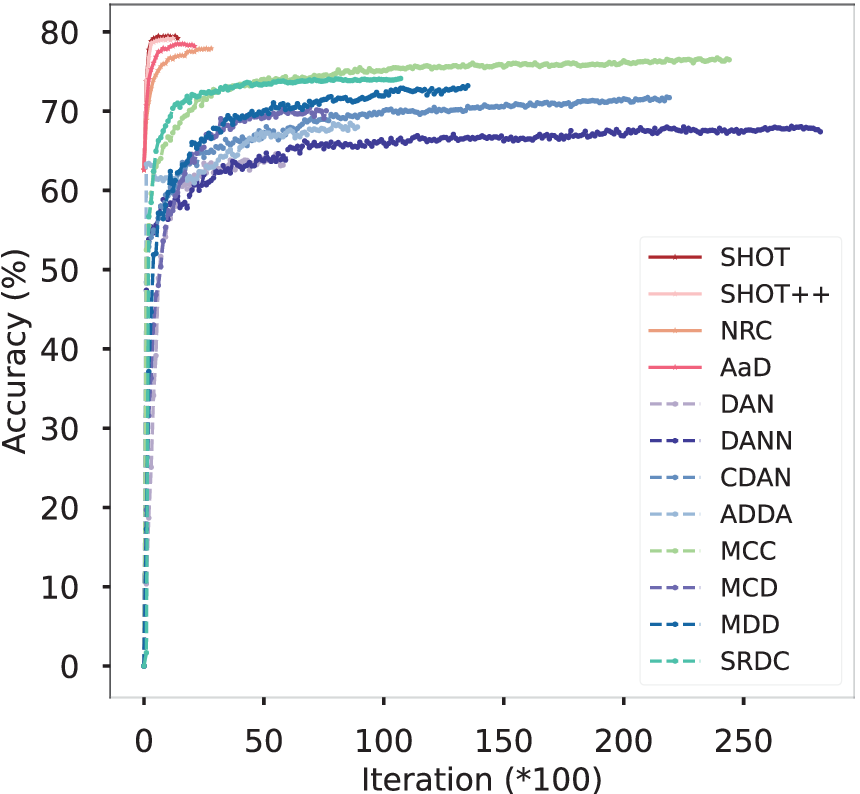}
\subcaption{Cl$\rightarrow$Pr}
\end{minipage}

\begin{minipage}[b]{.245\linewidth}
\centering
\includegraphics[height=4cm,width=4.4cm]{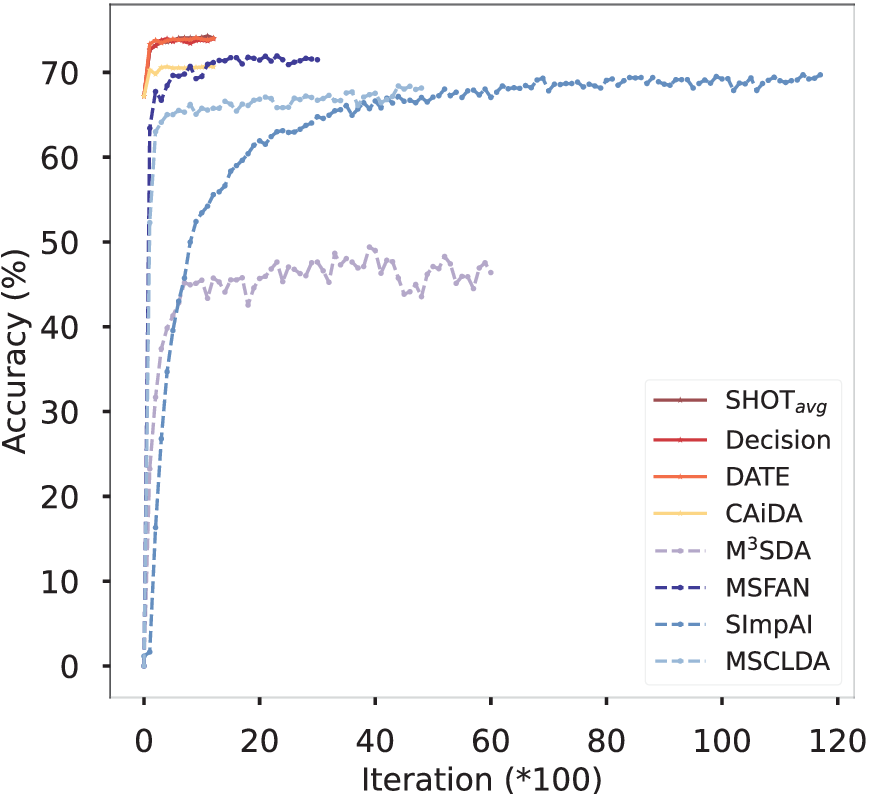}
\subcaption{Ar}
\end{minipage}
\begin{minipage}[b]{.245\linewidth}
\centering
\includegraphics[height=4cm,width=4.4cm]{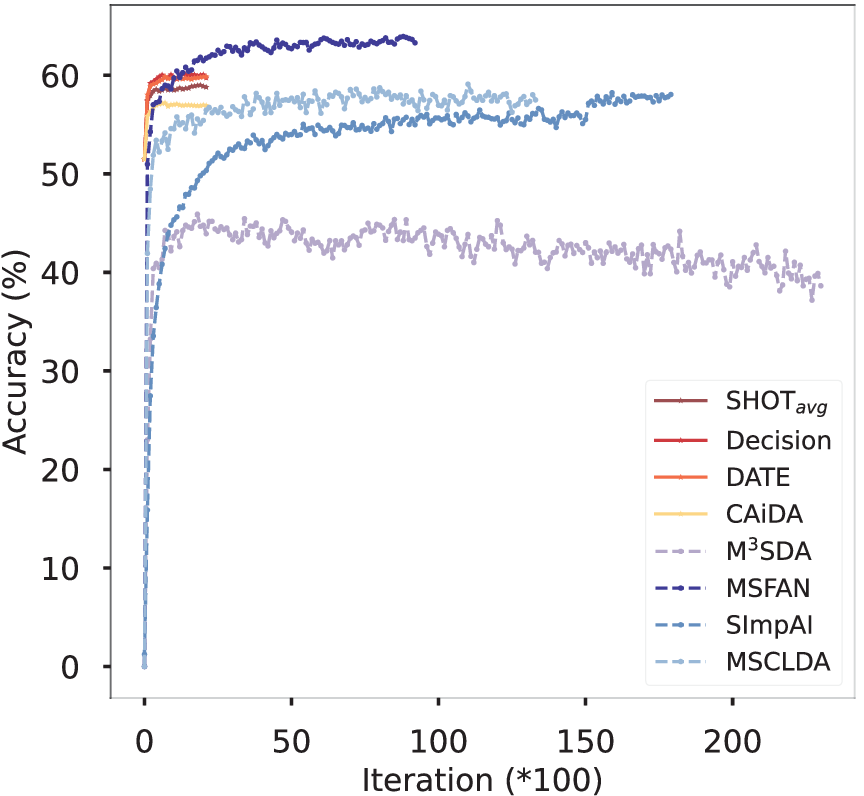}
\subcaption{Cl}
\end{minipage}
\begin{minipage}[b]{.245\linewidth}
\centering
\includegraphics[height=4cm,width=4.4cm]{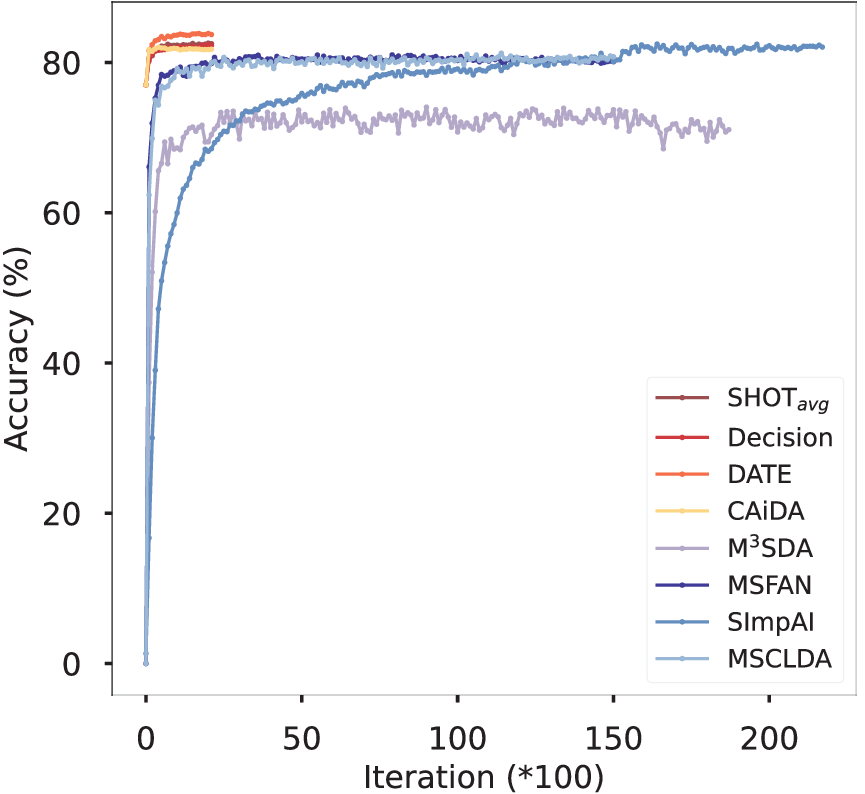}
\subcaption{Pr}
\end{minipage}
\begin{minipage}[b]{.245\linewidth}
\centering
\includegraphics[height=4cm,width=4.4cm]{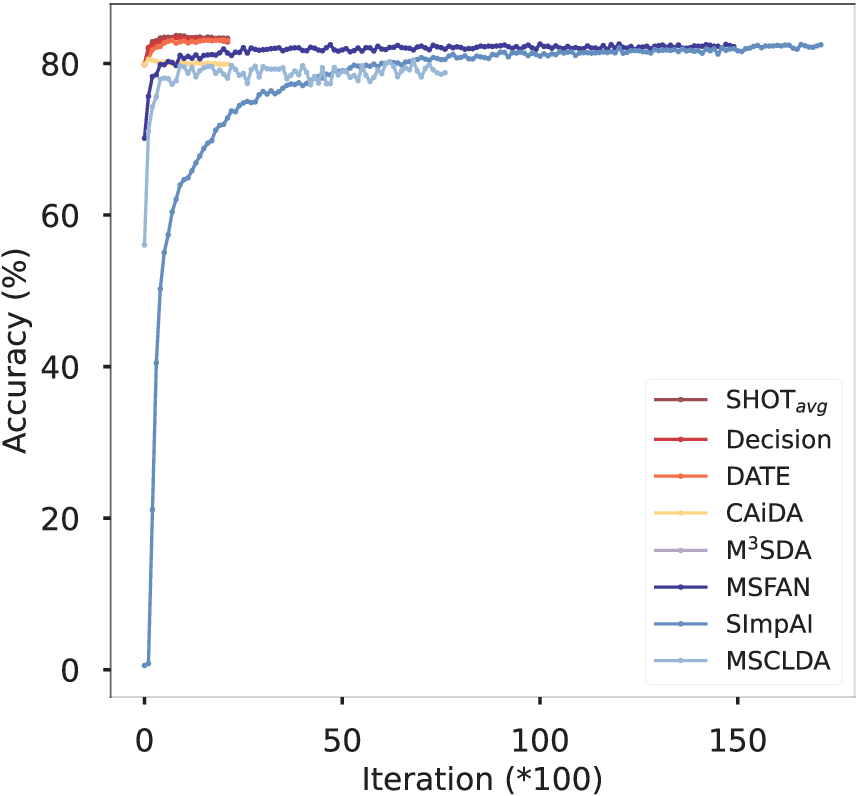}
\subcaption{Re}
\end{minipage}
\caption{Time analysis of UDA and SFDA methods in single-source and multi-source settings on the Office-Home dataset using RN50. We applied classical SFDA and UDA methods to four single-source transfer tasks (Ar$\rightarrow$Cl, Ar$\rightarrow$Pr, Cl$\rightarrow$Ar, Cl$\rightarrow$Pr ) and four multi-source transfer tasks on Office-Home (Ar, Cl, Pr, Re). SFDA methods (solid lines) consistently converge within approximately 200 iterations, while UDA methods (dashed lines) require at least five times more iterations due to the need to retrain with source data. These results demonstrate SFDA's superior efficiency in both single-source and multi-source settings.}
\label{Fig_time_ana_intro}
\end{figure*}


\section{Introduction}

\begin{figure*}[h]
\begin{minipage}[b]{.495\linewidth}
\centering
\includegraphics[height=3.5cm,width=8cm]{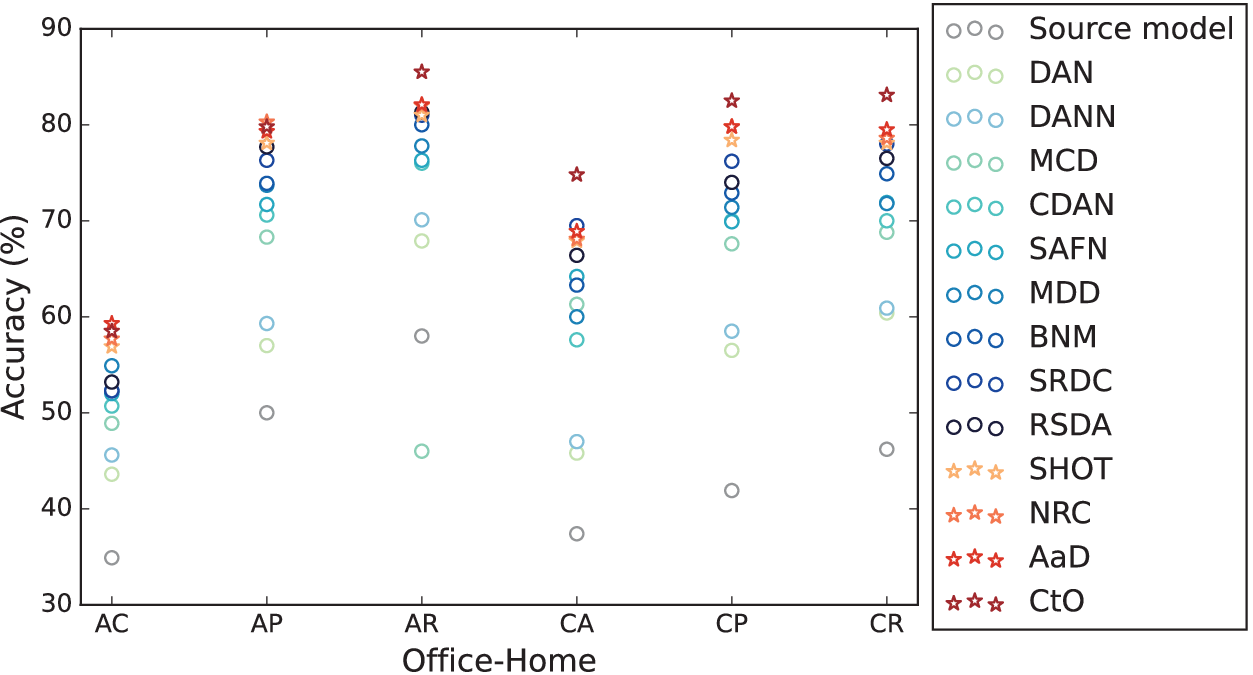}
\subcaption{Transfer tasks in Office-Home}
\end{minipage}
\begin{minipage}[b]{.495\linewidth}
\centering
\includegraphics[height=3.5cm,width=8cm]{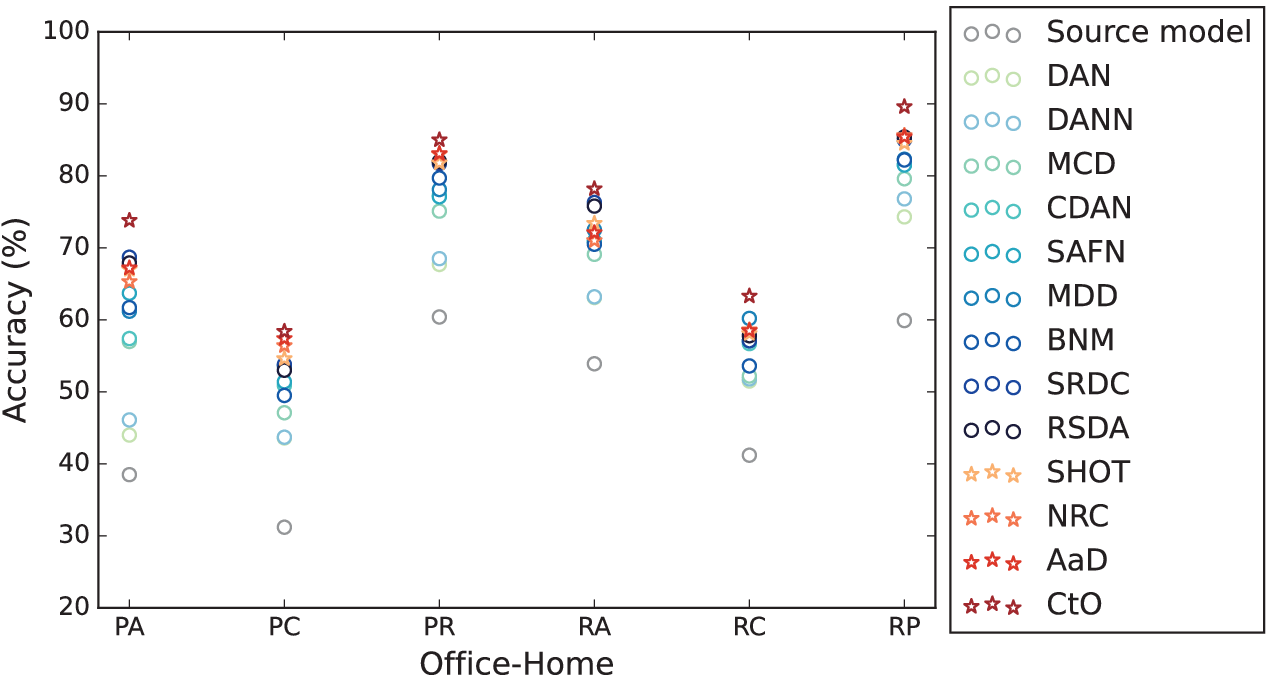}
\subcaption{Transfer tasks in Office-Home}
\end{minipage}

\begin{minipage}[b]{.495\linewidth}
\centering
\includegraphics[height=3.5cm,width=8cm]{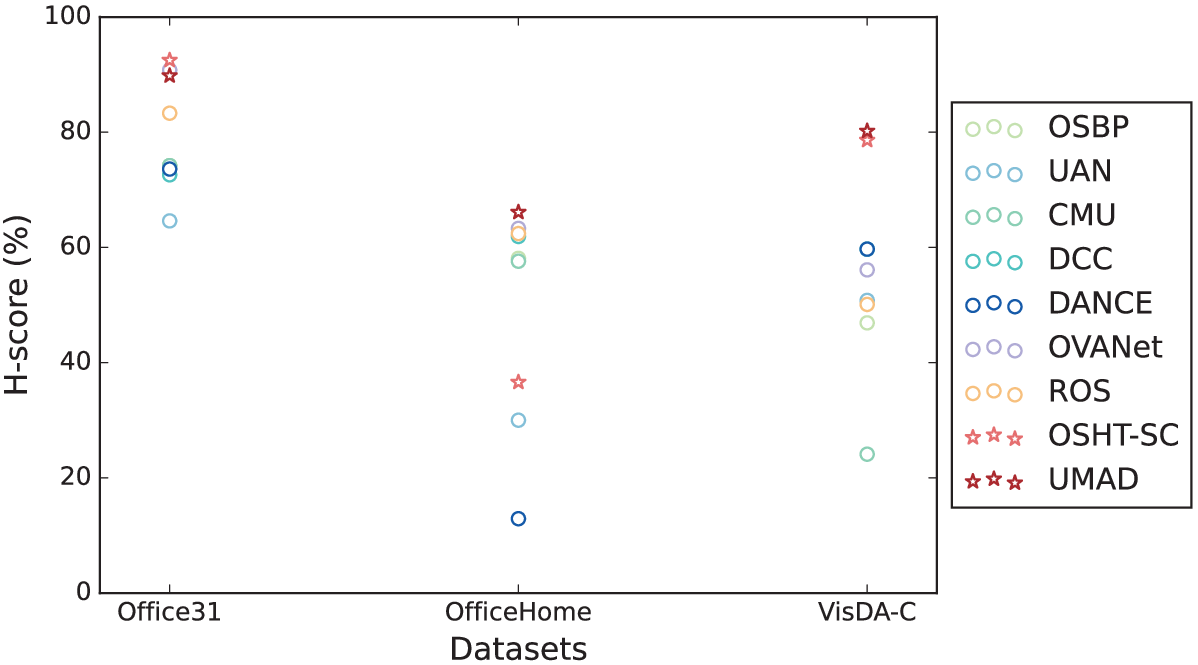}
\subcaption{Open set domain adaptation}
\end{minipage}
\begin{minipage}[b]{.495\linewidth}
\centering
\includegraphics[height=3.5cm,width=8cm]{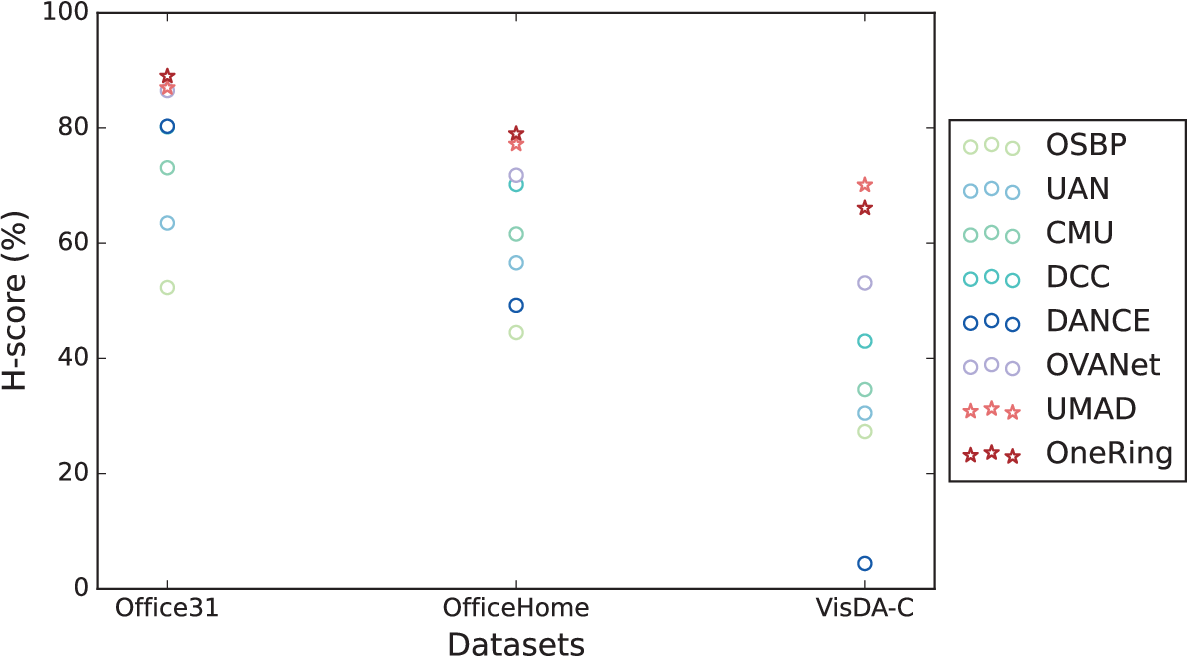}
\subcaption{Open partial domain adaptation}
\end{minipage}

\begin{minipage}[b]{.495\linewidth}
\centering
\includegraphics[height=3.5cm,width=8cm]{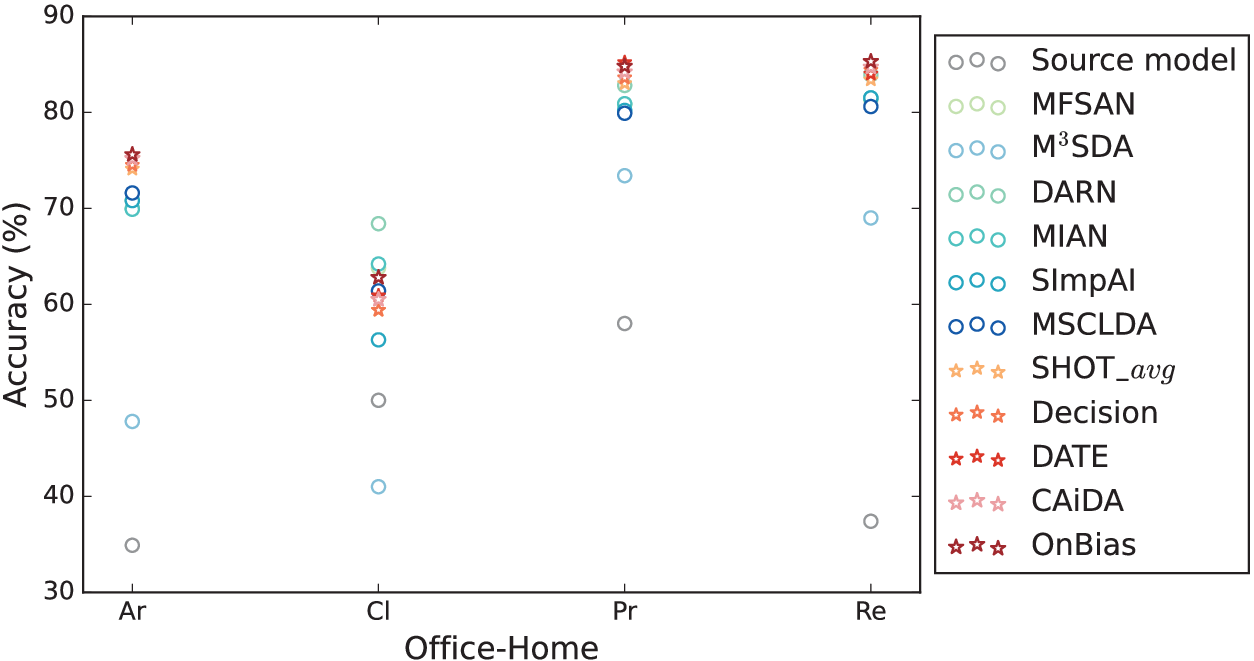}
\subcaption{Office-Home}
\end{minipage}
\begin{minipage}[b]{.495\linewidth}
\centering
\includegraphics[height=3.5cm,width=8cm]{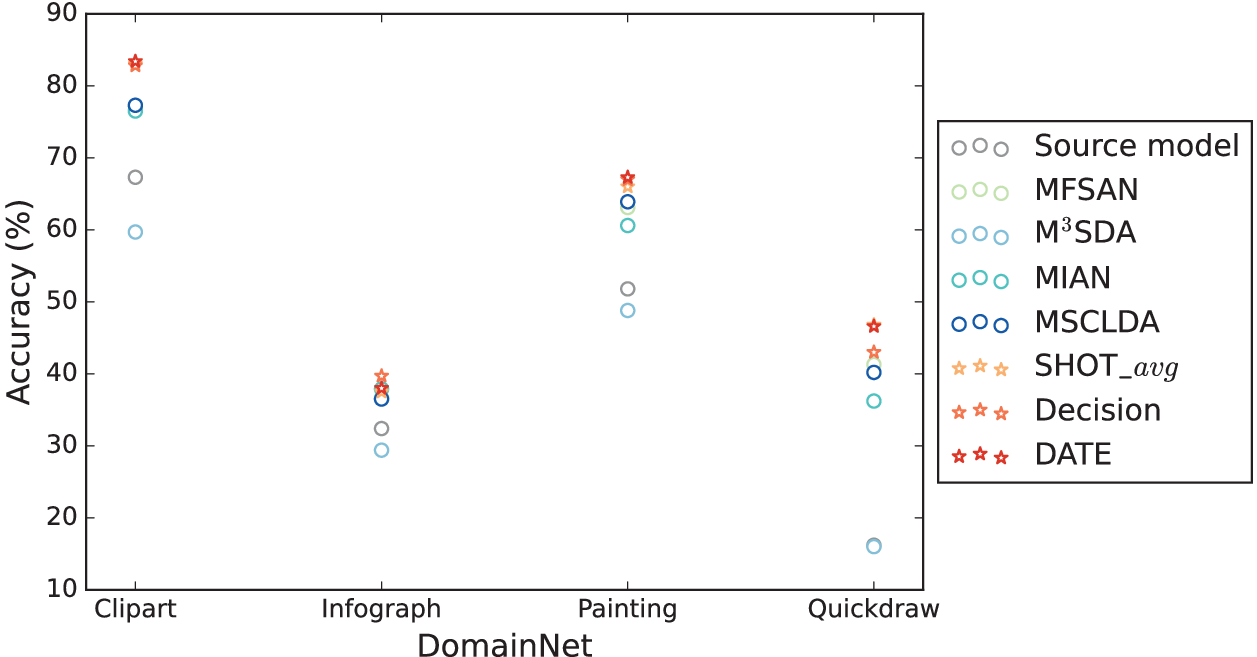}
\subcaption{DomainNet(50)}
\end{minipage}

\begin{minipage}[b]{.495\linewidth}
\centering
\includegraphics[height=3.5cm,width=8cm]{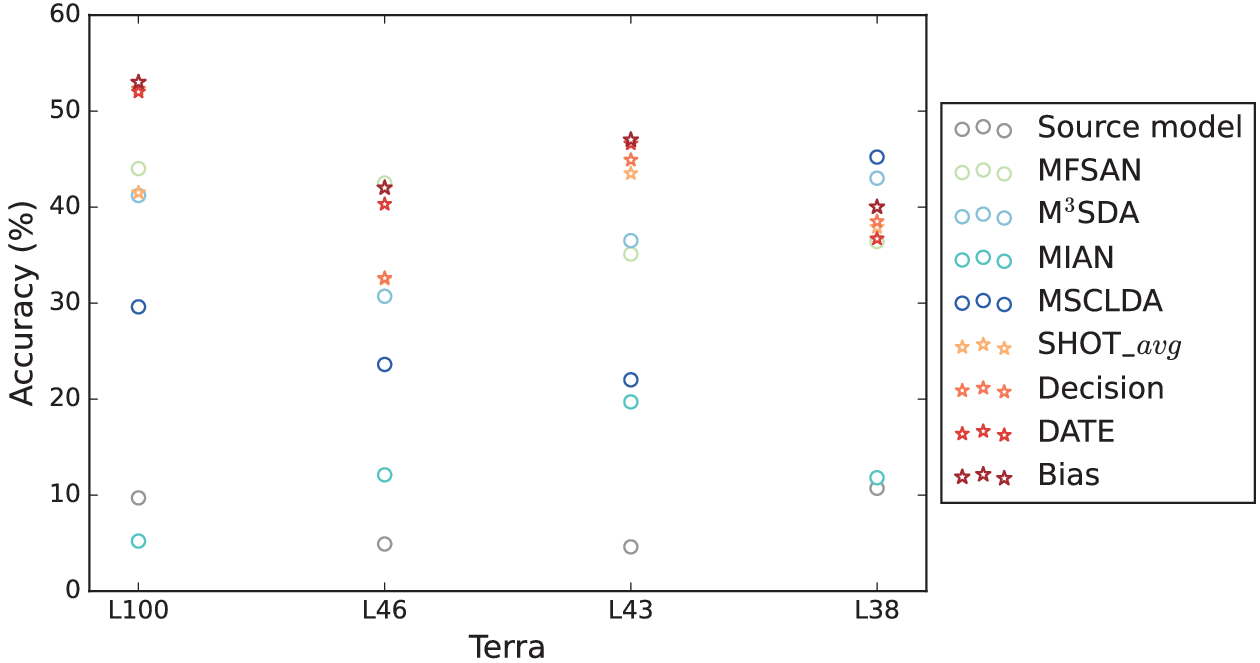}
\subcaption{Terra}
\end{minipage}
\begin{minipage}[b]{.495\linewidth}
\centering
\includegraphics[height=3.5cm,width=8cm]{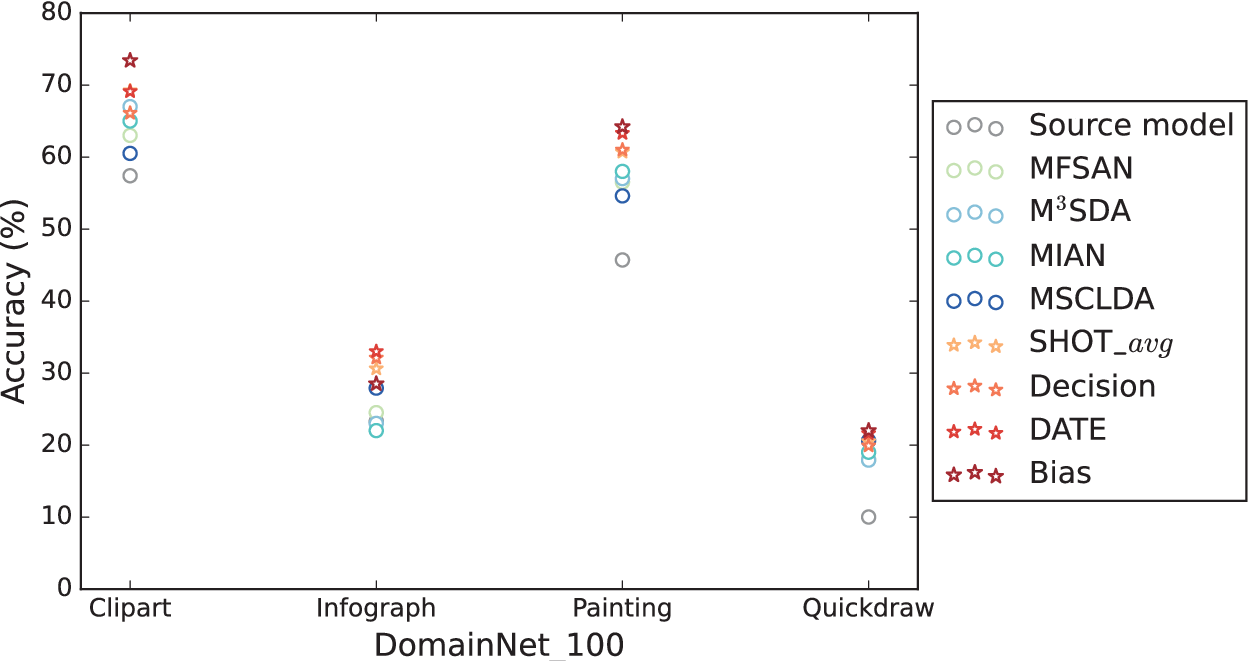}
\subcaption{DomainNet(100)}
\end{minipage}
\caption{Performance analysis with RN50.  (a) and (b) depict single-source domain adaptation in Office-Home to tackle covariate shift. (c) and (d) illustrate single-source domain adaptation in Office-31, Office-Home, and VisDA-C to address label shift. Open partial domain adaptation denotes the existence of novel classes in both source and target domains, as well as shared classes. Open set domain adaptation reveals the emergence of novel classes in the target domain that are absent in the source domain. 
(e), (f), (g), and (h) showcase multi-source domain adaptation in Office-Home, DomainNet (50 classes), Terra, and DomainNet(100 classes). Notably, SFDA methods are denoted by stars while UDA methods are represented by circles. 
Moreover, we reproduce the DomainNet and Terra results in (f), (g) and (h) and record other results as per the original publication.
}
\label{Fig_performance_intro}
\end{figure*}

In scenarios where the distributions of the training and testing datasets are consistent, contemporary machine learning algorithms excel in various applications, such as image recognition \cite{ferenc2020deep}, speech recognition \cite{zhang2015digital}, and data mining \cite{li2015twitter}. These algorithms rely on the assumption of a stable environment to achieve high accuracy and reliability. However, when this assumption of distributional stability is violated, their performance deteriorates sharply, revealing a critical vulnerability \cite{long2017conditional,long2015learning}. This lack of robustness in non-stationary environments highlights the necessity for developing more adaptable and resilient machine learning methods capable of maintaining performance despite distributional shifts.

Unsupervised Domain Adaptation (UDA) effectively addresses distribution discrepancies between training (source) and testing (target) datasets. Building on Ben-David's theoretical framework \cite{ben2007analysis,ben2010theory}, UDA methods primarily align domain distributions to enhance target performance. Common approaches include minimizing discrepancy loss \cite{chen2020homm,peng2019moment,zellinger2019robust,nguyen2024class,qiang2021robust} and adversarial learning, which trains a domain discriminator alongside a feature extractor to create domain-invariant features \cite{long2017conditional,saito2017adversarial,tzeng2017adversarial,Tang_2020_CVPR}. Recently, Source-Free Domain Adaptation (SFDA) has emerged as a powerful paradigm for managing distribution shifts while ensuring data privacy. SFDA adapts models to target domains without accessing source data by leveraging pre-trained source models. It typically employs techniques such as self-supervised learning \cite{liangjian,liang2021source,kundu2022balancing,DBLP:journals/jcst/TianMZPX21} or contrastive learning \cite{yang2021exploiting,yang2021generalized,yang2022attracting,yang2022one} for adaptation.

Determining whether UDA or SFDA is superior is a crucial and urgent question in domain adaptation, as these paradigms have not been directly compared, creating a gap in understanding their practical strengths and limitations. Without this comparison, selecting the optimal approach for adaptable and privacy-sensitive systems is challenging, affecting system efficiency, data privacy, and adaptability in environments with restricted access to source data. Clarifying the relative advantages of UDA and SFDA is thus both theoretically important and has immediate real-world implications.

This paper presents a comprehensive analysis, combining theoretical insights and extensive experiments, to determine whether UDA or SFDA is better suited for practical domain adaptation tasks. Drawing on predictive coding theory \cite{rao1999predictive, spratling2017review}, which posits that the brain adapts to new information using existing models, we argue that SFDA aligns with this efficient, adaptive theory. SFDA leverages pre-trained source models to interpret new data, enhancing adaptability and reducing cognitive load without accessing source data. Our experiments support SFDA's theoretical advantages, demonstrating that across multiple benchmark datasets, SFDA outperforms UDA in time efficiency, memory usage, learning objectives, resistance to negative transfer, and robustness against overfitting (see Fig. \ref{Fig_time_ana_intro} and \ref{Fig_performance_intro}). Specifically, SFDA converges within approximately 200 iterations, while UDA methods require 1,000 to 5,000 iterations to achieve similar performance, especially when significant distributional discrepancies exist, as seen in the Terra and DomainNet datasets. These findings validate SFDA's practical relevance and effectiveness in scenarios with substantial domain shifts.

Additionally, we introduce the data-model fusion scenario for the first time, addressing the complex requirements of multi-party data sharing in real-world applications. In fields like healthcare, public institutions may provide raw data while private entities can only share pre-trained models due to privacy restrictions. Traditional UDA and SFDA methods are limited as they rely solely on source data or source models, respectively, preventing full utilization of resources across multiple parties. To overcome these challenges, we propose a novel data-model fusion strategy that builds on SFDA’s strengths and incorporates an advanced weight estimation method within a multi-source-free framework. This approach trains source models on available data and assigns weights based on their performance on other source domains, effectively evaluating their ability to extract domain-invariant features. Our experiments show that this method significantly outperforms existing Multi-UDA and Multi-SFDA techniques, demonstrating its effectiveness in complex data-model fusion scenarios. This strategy enhances model performance in target domains where data and models are shared among entities with diverse data-sharing constraints, providing a practical solution for real-world applications.

The contributions of this paper are summarized as follows:

\begin{itemize}
\item We present a detailed analysis, grounded in predictive coding theory, that illustrates why 
 SFDA methods align more effectively with real-world adaptation requirements than UDA methods. This analysis highlights the adaptability and efficiency of SFDA, particularly in cases with substantial domain distribution discrepancies. 

\item We conduct extensive experiments on benchmark datasets demonstrating that SFDA methods surpass UDA methods in time efficiency, storage capacity, and in preventing negative transfer and overfitting.

\item We introduce a practical data-model fusion scenario for the first time and propose a weight estimation strategy within a multi-source-free framework to optimize model adaptation in such scenarios. Our approach outperforms existing Multi-UDA and Multi-SFDA methods, addressing their limitations when data and models are variably shared across stakeholders.

\end{itemize}

\section{Related Work}
\subsection{Unsupervised Domain Adaptation}
\subsubsection{Single-source unsupervised domain adaptation (UDA)} It transfers knowledge from a labeled source domain to an unlabeled target domain, addressing domain shifts by aligning distributions. Foundational theories \cite{ben2007analysis, ben2010theory} emphasize distribution alignment as a key to reducing discrepancies. Discrepancy-based approaches, such as DAN \cite{long2015learning} and JAN \cite{long2017deep}, employ Maximum Mean Discrepancy (MMD) to align domain distributions, while HoMM \cite{chen2020homm} matches higher-order statistical moments to minimize discrepancies. More advanced methods like MDD \cite{zhang2019bridging} and MMFND \cite{xu2019larger} extend these ideas with tailored loss functions for improved feature alignment. Adversarial methods such as DANN \cite{ganin2015unsupervised} introduce domain discriminators to promote domain-invariant features, while extensions like CDAN \cite{long2017conditional} integrate classifier predictions for conditional alignment. Recent techniques, such as SRDC \cite{Tang_2020_CVPR}, explore target domain structures with source-guided regularization. Despite their success, these methods often struggle in open environments where source data is inaccessible.

\subsubsection{Multi-source unsupervised domain adaptation (MUDA)} It extends UDA by leveraging multiple labeled source domains to adapt to an unlabeled target domain, aiming to minimize discrepancies between the target and each source as well as among the sources. Methods such as DCTN \cite{xu2018deep} use domain discriminator scores for source-target similarity, while MFSAN \cite{zhu2019aligning} employs maximum mean discrepancy for domain-invariant features and consistent classifiers. M$^3$SDA \cite{peng2019moment} and STEM \cite{nguyen2021stem} align distributions using MMD loss and teacher-student frameworks. CMSS \cite{yang2020curriculum} and MIAN \cite{park2021information} select target-like sources via additional discriminators and information-theoretic approaches. SImpAI \cite{venkat2020your} progressively adapts from easy to hard samples, while MDDA \cite{zhao2020multi}, MSCLDA \cite{li2021multi}, DRAN \cite{wen2020domain}, and TMDA \cite{wang2019tmda} use Wasserstein distance to select and align similar source samples. However, like single-source UDA, MUDA requires access to source data, posing challenges in privacy-sensitive or restricted environments.

\subsection{Source Free Domain Adaptation}
\subsubsection{Single-source-free domain adaptation (SFDA)}  It shifts the focus to adapting from a pre-trained source model to an unsupervised target domain, eliminating the need for source data. Self-training methods such as SHOT \cite{liangjian}, HCL \cite{huang2021model}, and NRC \cite{yang2021exploiting} refine target predictions using clustering and self-supervised learning. Adversarial extensions like A$^2$Net \cite{xia2021adaptive} iteratively improve adaptation, while AaD \cite{yang2022attracting} combines adversarial and self-training strategies. Data generation approaches, such as MA \cite{li2020model} and SDDA \cite{kurmi2021domain}, synthesize source-like data or generate pseudo-labels for target data. While effective, data generation methods often incur high computational costs. SFDA techniques have demonstrated significant improvements across benchmarks and medical applications \cite{guan2021domain,ye2022alleviating,yang2022source}, emphasizing their scalability and privacy compliance.

\subsubsection{Multi-source-free domain adaptation (MSFDA)} It leverages multiple source models to adapt to a target domain without accessing source data, enabling efficient knowledge transfer from diverse domains. MSFDA methods focus on two key aspects: assigning weights to source models based on their impact on the target and adapting single-source FDA techniques for effective adaptation. For weighting, Decision \cite{ahmed2021unsupervised} uses dynamic learning based on target performance, CAiDA \cite{dong2021confident} applies confident adaptive learning based on model reliability, MSFUDA \cite{pei2024evidential} employs evidential reasoning to iteratively update weights, and OnBias \cite{shen2023balancing} balances source model biases through dynamic reweighting. Alternatively, MSFDA \cite{li2023target} assesses static transferability, MFRA \cite{pei2024evidential} uses fixed ranking based on target discriminability, and DATE \cite{han2023discriminability} optimizes weights by evaluating static discriminability. Additionally, MSFDA integrates advanced SFDA techniques \cite{liangjian,yang2021exploiting} to enhance knowledge transfer effectively.

Although UDA and SFDA use different methods, their growing differences highlight the need for a thorough comparison. Without such an evaluation, researchers might overlook important trade-offs and limitations, hindering the development of more effective domain adaptation strategies. This study suggests that SFDA, with its scalability and privacy-preserving features, is better suited to address real-world challenges. By carefully comparing these approaches, this work highlights their strengths and weaknesses, demonstrating why SFDA is a more robust and practical choice for advancing domain adaptation.

\section{Theory, Experiments and Analysis: SFDA methods are better than UDA methods}

This section starts by detailing the preliminaries and experimental setup, including the datasets and baselines employed. It then establishes, through both predictive coding theory and empirical results, that SFDA is more suitable and performs better in real-world scenarios compared to UDA. Finally, it provides an in-depth analysis from multiple dimensions, such as time efficiency, storage capacity, learning objectives, and negative transfer, to elucidate the reasons why SFDA outperforms UDA.

\subsection{Preliminaries and Experimental setup} 

In this paper, we denote $D_s^1$ = $\{x_i^{s_1}, y_i^{s_1}\}_{i=1}^{n_{s_1}}$, $\cdots$, $D_s^m$ = $\{x_i^{s_m}, y_i^{s_m}\}_{i=1}^{n_{s_m}}$ as the source domains, where $m$ denotes the number of source domains and $n_{s_m}$ represents the labeled samples in the corresponding source domain. It is important to note that $y$  $\in$ $\mathcal{Y}$  $\in$ $\mathcal{R}^K$  idenotes the one-hot ground-truth label, with $K$ being the total number of classes within the label set. Furthermore, $D_t$ = $\{x_i^t\}_{i=1}^{n_t}$ denotes the unsupervised target domain comprising of $n_t$ unlabeled samples, which share the same underlying label set as the source data. \textbf{The objective of UDA} methods is to transfer the knowledge from the source domain(s) to the target domain.  In source-free domain adaptation (SFDA) scenario, we have access to the source model(s) $M(s)$, which has been previously trained on the source domain(s) in a supervised manner using cross-entropy loss. Specifically, $M^i$ consists of a feature extractor and a linear classifier.  \textbf{The objective of SFDA} methods is to facilitate the transfer of knowledge from the designated source model(s) to the unsupervised target domain.

\subsubsection{Datasets}
We adopt Five benchmark datasets, including Office-31 \cite{saenko2010adapting}, Office-Home \cite{venkateswara2017deep}, and VLCS \cite{fang2013unbiased}  with small distribution differences, as well as DomainNet \cite{peng2019moment} and TerraIncognita \cite{beery2018recognition} with significant distribution variations.

\textbf{Office-31} serves as a conventional benchmark for domain adaptation, sourcing images from three distinct domains: Amazon (\textbf{A}), Webcam (\textbf{W}), and DSLR (\textbf{D}). These three domains collectively encompass 31 classes, with sample sizes of 2817, 795, and 498, respectively.

\textbf{Office-Home} is a demanding medium-sized benchmark, comprising four distinct domains: Artistic images (\textbf{Ar}), Clip Art (\textbf{Cl}), Product images (\textbf{Pr}), and Real-world images (\textbf{Rw}). Each domain comprises a total of 65 every object categories.

\textbf{VLCS}  is an amalgamation of Caltech101, LabelMe, PASCAL, and SUN09, each individually designed for object recognition, yet incorporating biases specific to their datasets. It comprises 10,729 examples from 5 classes.

\textbf{DomainNet} encompasses six discrete domains, 345 categories, and approximately 0.6 million images, with domains including Clipart, Infograph, Painting, Quickdraw, Real, and Sketch. To manage the broad scope of domains and sample sizes, we opt to experiment solely with the initial four domains (Clipart (\textbf{C}), Infograph (\textbf{I}), Painting (\textbf{P}), Quickdraw (\textbf{Q})) and the top 30 classes and 100 classes within each domain in our experiments.

\textbf{TerraIncognita} (Terra) comprises photographs of wild animals captured by cameras in various locations. Utilizing datasets $d$ $\in$ \{\textbf{L100}, \textbf{L38}, \textbf{L43}, \textbf{L46}\} as outlined in \cite{gulrajani2020search}, the dataset consists of 24,788 examples across 10 classes.

\subsubsection{Baselines} In single-source and multi-source settings, we select classical methods as follows:

For single-source setting, UDA methods include ResNet-50 \cite{he2016deep}, DAN \cite{long2015learning}, DANN \cite{ganin2016domain}, MCD \cite{saito2018maximum}, ADDA \cite{tzeng2017adversarial}, MDD \cite{zhang2019bridging}, MCC \cite{jin2020minimum}, and SRDC \cite{Tang_2020_CVPR}. SFDA methods consist of source model, SHOT \cite{liangjian}, SHOT$_{plus}$ \cite{liang2021source}, NRC \cite{yang2021exploiting}, and AaD \cite{yang2022attracting}. Other methods like UTR \cite{pei2023uncertainty} and CPGA \cite{qiu2021source} are challenging to reproduce, hence they are not compared.

For multi-source setting, we utilize classical MUDA methods MFSAN \cite{zhu2019aligning}, M$^3$SDA \cite{peng2019moment}, MIAN \cite{park2021information}, STEM \cite{nguyen2021stem}, and MSCLDA \cite{li2021multi}. Additionally, we employ MSFDA methods such as SHOT$_{Avg }$\cite{liangjian}, Decision \cite{ahmed2021unsupervised}, DATE \cite{han2023discriminability}, CAiDA  \cite{dong2021confident}, and onBias \cite{shen2023balancing}. Similarly, other methods like LtC-MSDA \cite{wang2020learning}, T-SVDnet \cite{li2021t},  MOST \cite{nguyen2021most}, TMDA \cite{wang2019tmda} are difficult to reproduce in our datasets using their open-source code, thus we do not compare them.

\subsubsection{Implementation details} Primary results are demonstrated using ResNet-50 as the backbone for all datasets. The network architecture mirrors established methods for unsupervised domain adaptation. In source-free domain adaptation, the network architecture aligns with SHOT \cite{liangjian}. Reproduction of existing UDA methods is conducted using open source code available at Tsinghua University's GitHub repository. Default parameters and settings of specific methods are applied. A code containing all reproduced datasets and methods in this study will be publicly shared upon the publication of the paper.

\subsection{Theory Analysis: Relationship Between Predictive Coding Theory and SFDA Paradigm}
Here, we delve deeper into the relationship between predictive coding theory and Source-Free Domain Adaptation (SFDA) methods. We first introduce predictive coding theory as a theoretical framework that mirrors real-world learning dynamics. We then contextualize SFDA within this framework, elucidating how it embodies predictive coding principles. Finally, we analyze the theoretical advantages of SFDA derived from predictive coding, highlighting its practical significance.

\textbf{Predictive Coding Theory: A Framework for Efficient Learning.}  
Predictive coding theory, rooted in cognitive science, offers a comprehensive explanation of how the brain processes information efficiently by generating predictions~\cite{spratling2017review}. According to this theory, the brain actively constructs internal models based on prior knowledge and experience, which it uses to predict incoming sensory inputs. When actual sensory inputs deviate from predictions, the resulting prediction errors are used to update and refine these models, improving the brain’s capacity to anticipate future inputs.

This theory is particularly relevant to real-world scenarios for several reasons. First, predictive coding emphasizes active prediction and error correction, allowing efficient and adaptive learning by iteratively addressing discrepancies between predictions and actual inputs. Second, it supports adaptability in dynamic environments by enabling models to evolve in response to changes, ensuring robustness and flexibility in the face of variability. Third, it fosters efficient resource utilization by focusing on unexpected inputs and disregarding redundant or familiar data, thereby minimizing computational costs and reducing cognitive load. The predictive coding process involves generating predictions based on accumulated knowledge, observing actual inputs from the environment, calculating prediction errors by quantifying discrepancies between predictions and observations, and refining the internal models to improve future accuracy. This iterative mechanism ensures efficiency and adaptability, key features that align closely with SFDA methods.

\textbf{Mapping Predictive Coding to SFDA Methods.}  
SFDA methods inherently embody predictive coding principles by leveraging pre-trained source models as internal representations of prior knowledge. These models predict features or labels for the target domain, analogous to how the brain generates predictions from internal models. Target domain data serve as new inputs, with discrepancies between source predictions and target features quantified using metrics such as entropy or pseudo-label confidence. These errors are used to update the model, typically through self-supervised or contrastive learning techniques, aligning the source model more closely with the target domain. This process mirrors predictive coding’s cycle of prediction, error correction, and model refinement, demonstrating SFDA’s adherence to this efficient and adaptive learning paradigm.

\textbf{Theoretical Advantages of SFDA Based on Predictive Coding.}  
Based on predictive coding theory, SFDA offers several theoretical advantages over UDA. Its adaptability to distribution shifts ensures robust performance in dynamic environments with significant variability. By focusing on prediction errors and leveraging pre-trained models, SFDA minimizes computational overhead, making it well-suited for resource-constrained settings. The isolation of target-specific features reduces the risk of negative transfer, particularly in cases of severe domain shifts. Furthermore, by eliminating the need for source data during adaptation, SFDA enhances data privacy and security, making it ideal for privacy-sensitive applications like healthcare. Finally, the iterative error correction inherent in predictive coding allows SFDA to converge more rapidly than UDA, as supported by our empirical results.

\textbf{Practical Significance.}  
By grounding SFDA in predictive coding theory, we provide a theoretical foundation for its practical advantages. SFDA’s ability to adapt rapidly, efficiently, and securely makes it a superior paradigm for scenarios with stringent privacy constraints, dynamic environments, or limited computational resources. This connection to predictive coding not only validates the effectiveness of SFDA but also inspires further exploration of predictive coding principles in machine learning paradigms.

\begin{figure*}[t]
\begin{minipage}[b]{.999\linewidth}
\centering
\includegraphics[height=3.5cm,width=18cm]{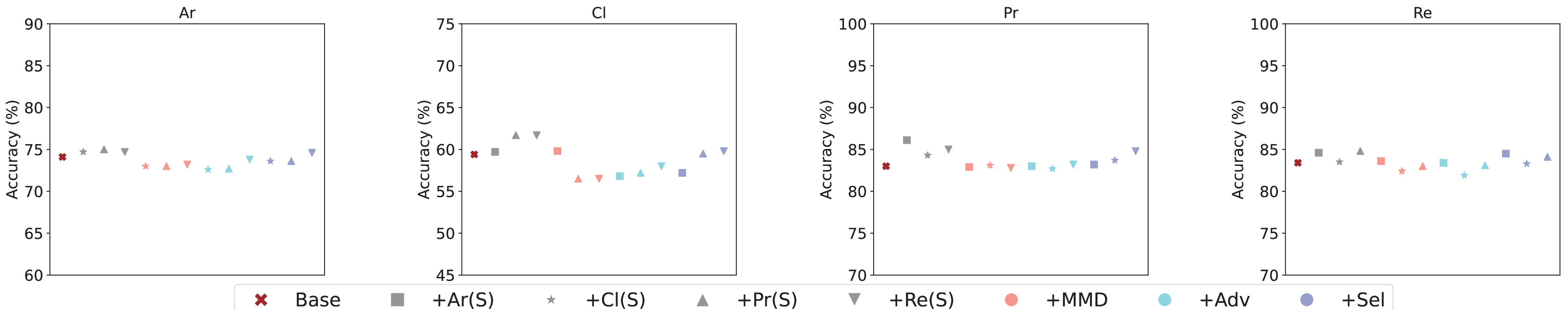}
\subcaption{SHOT$_{Avg}$ \cite{liangjian}}
\end{minipage}
\begin{minipage}[b]{.999\linewidth}
\centering
\includegraphics[height=3.5cm,width=18cm]{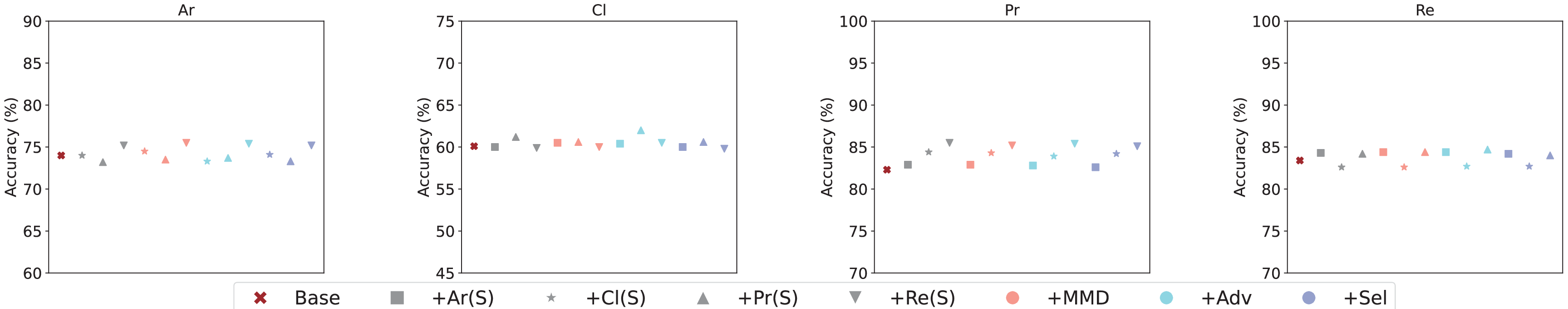}
\subcaption{Decision \cite{ahmed2021unsupervised}}
\end{minipage}

\begin{minipage}[b]{.999\linewidth}
\centering
\includegraphics[height=3.5cm,width=18cm]{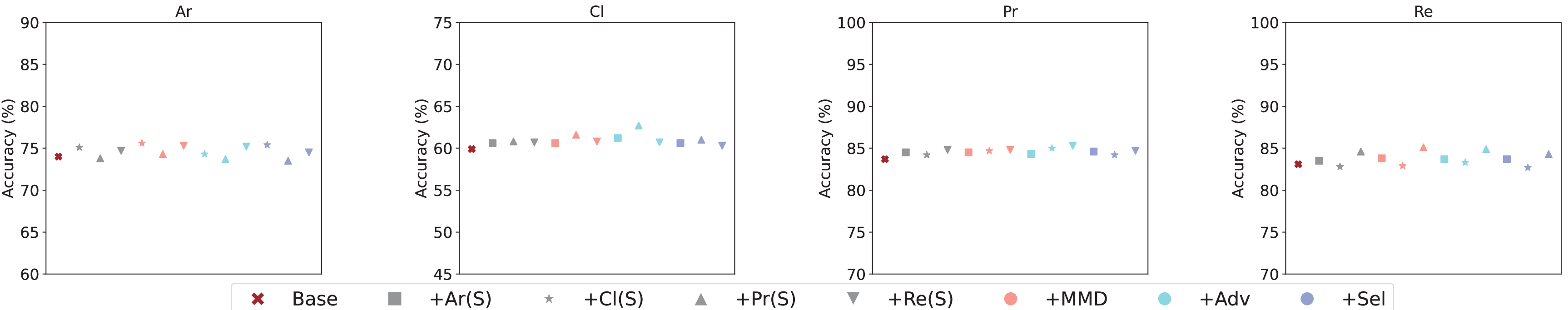}
\subcaption{Date \cite{han2023discriminability}}
\end{minipage}
\caption{Comparison of SFDA and UDA learning objectives on the Office-Home dataset: The `Base' method represents SFDA, with (a), (b), and (c) corresponding to SHOT, Decision, and Date. Gray images indicate adding labeled source data to `Base' as supervision. UDA methods are depicted with circles (MMD, Adv, Sel), while orange boxes and blue stars represent supervisory signals `Ar' and `Cl.' The figure demonstrates that for minor distribution shifts, adding source signals results in negligible improvement, typically under one percentage point.}
\label{Fig_performance_shot_compare}
\end{figure*}

\subsection{Experimental Analysis}

We further explore why SFDA methods outperform UDA, focusing on factors such as time efficiency, storage capacity, learning objectives, negative transfer, and overfitting.

\subsubsection{Time analysis and storage capacity analysis} Using single-source SFDA as an example, where only a pre-trained source model is provided, SFDA focuses solely on the adaptation time  ($T(A)$) within the target domain. In contrast, UDA requires both retraining time ($T(RA)$) in the source domain and adaptation time (which may overlap). Due to the need for retraining from scratch, UDA generally takes significantly longer than SFDA:
\begin{equation}  
\begin{aligned}  
T(A) \ll T(RA)\,.
\end{aligned} 
\label{Eq_uda_time}
\end{equation}
As the number of analogous target domains ($\alpha$) increases, the advantage of SFDA becomes even more evident. Unlike UDA, SFDA eliminates the need for repeated retraining of source data, making it highly scalable:

\begin{equation}  
\begin{aligned}  
T(\alpha A) \ll T(\alpha RA)\,.
\end{aligned} 
\label{Eq_muda_time}
\end{equation}
In multi-source domain adaptation, multi-SFDA is also faster than multi-UDA, as the latter requires retraining all data from multiple source domains.

Regarding storage, studies by Liang \cite{liangjian} highlight a significant difference in storage requirements between the source model and source data, with the source model requiring at least 40 orders of magnitude less storage. This characteristic makes SFDA particularly suitable for deployment in storage-constrained environments, such as mobile devices.

\begin{figure*}[t]
\begin{minipage}[b]{.999\linewidth}
\centering
\includegraphics[height=3.5cm,width=18cm]{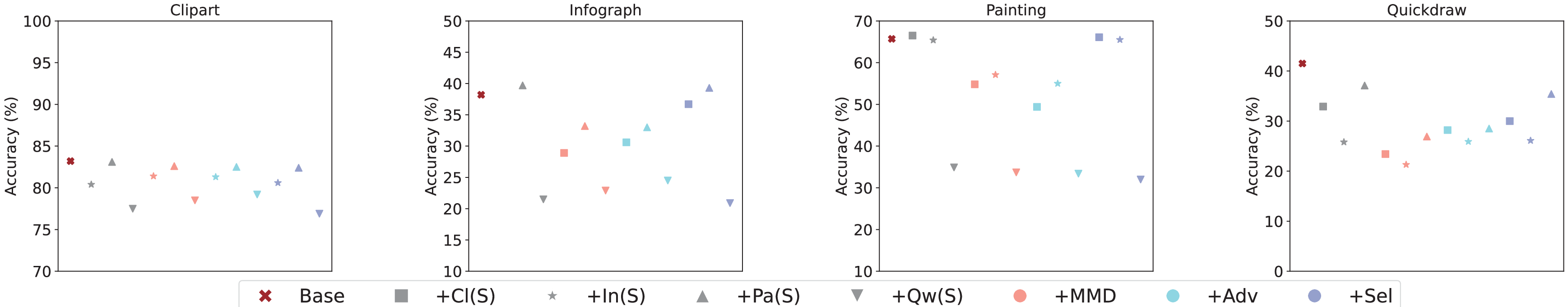}
\subcaption{SHOT$_{Avg}$ \cite{liangjian}}
\end{minipage}
\begin{minipage}[b]{.999\linewidth}
\centering
\includegraphics[height=3.5cm,width=18cm]{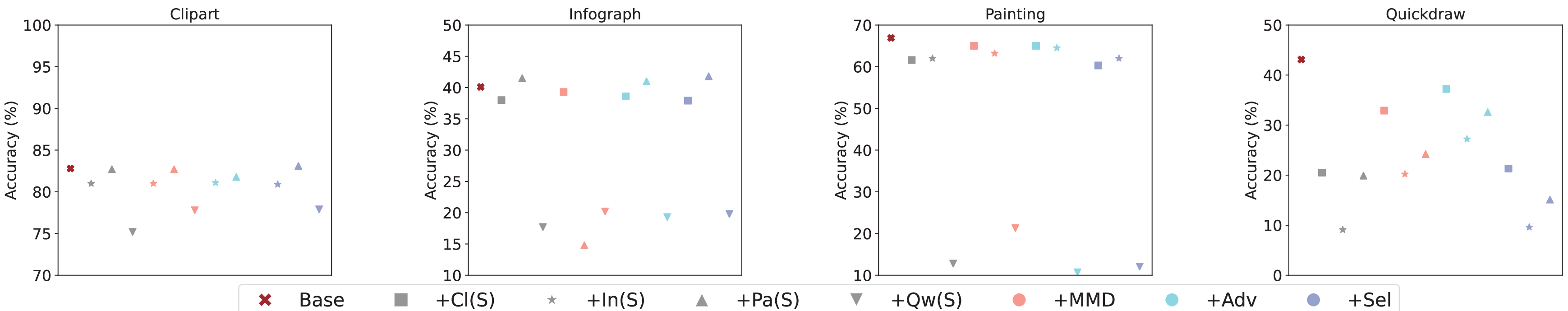}
\subcaption{Decision \cite{ahmed2021unsupervised}}
\end{minipage}

\begin{minipage}[b]{.999\linewidth}
\centering
\includegraphics[height=3.5cm,width=18cm]{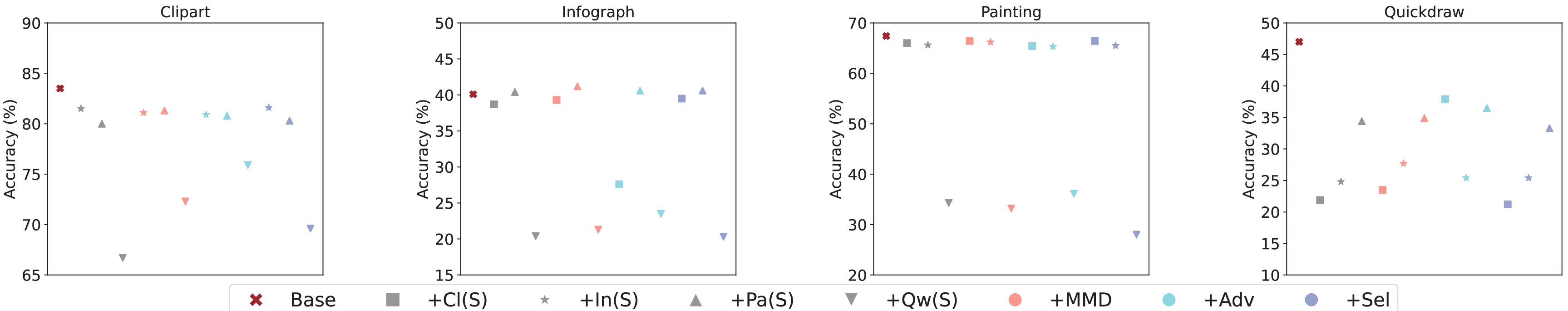}
\subcaption{Date \cite{han2023discriminability}}
\end{minipage}
\caption{Contrast between the learning objectives of SFDA methods and UDA methods on the DomainNet dataset. The other descriptions are same as the Fig .\ref{Fig_performance_shot_compare}. Observing the results depicted in the figure, it is evident that datasets with large distribution gaps, such as DomainNet, may experience notable performance degradation in specific transfer tasks despite the inclusion of source domain signals. The utilization of traditional UDA techniques like MMD, Adv, Sel does not completely mitigate the adverse effects of source data on target domain learning. }
\label{Fig_performance_dn_compare}
\end{figure*}

\begin{figure*}[t]
\begin{minipage}[b]{.999\linewidth}
\centering
\includegraphics[height=3.5cm,width=18cm]{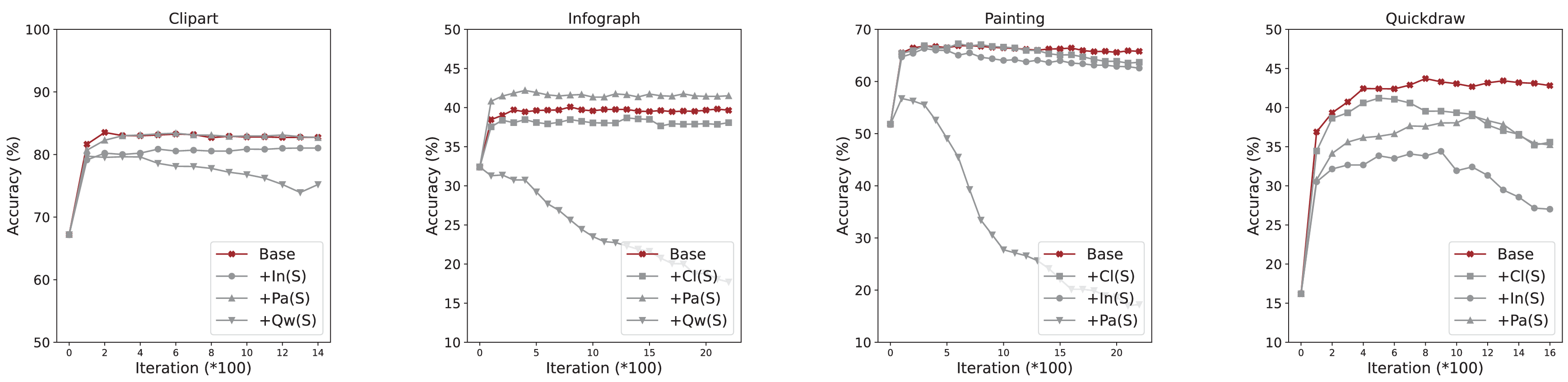}
\subcaption{Decision \cite{ahmed2021unsupervised}}
\end{minipage}
\begin{minipage}[b]{.999\linewidth}
\centering
\includegraphics[height=3.5cm,width=18cm]{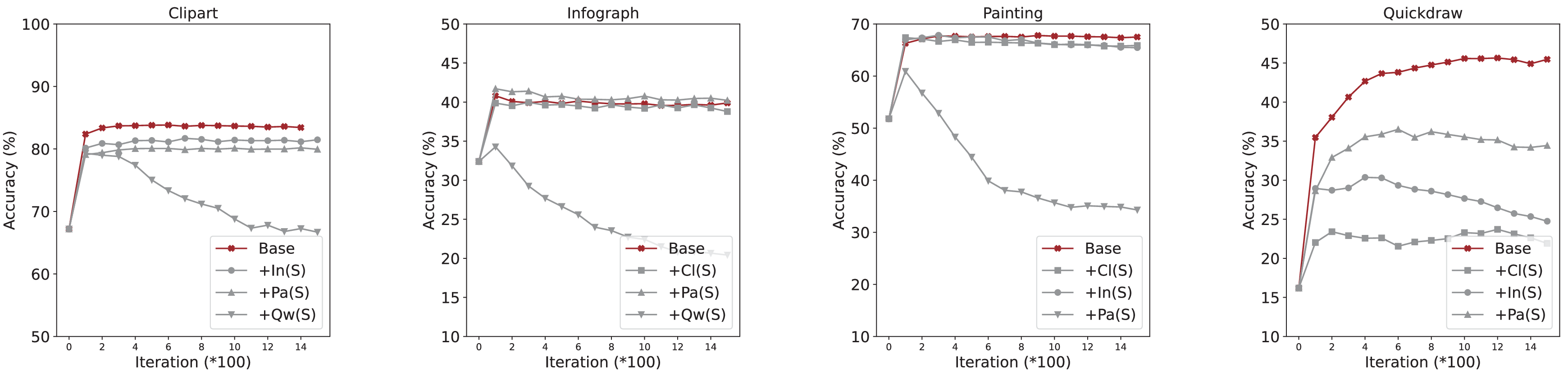}
\subcaption{Date \cite{han2023discriminability}}
\end{minipage}
\caption{Analysis of negative transfer of SFDA methods and UDA methods on the DomainNet dataset.  `Base' indicates the use of MSFDA algorithms, while grey signifies algorithms that integrate specific source data for supervised learning based on the Base. As training advances, a substantial disparity in distribution between the source and target domains leads to an increasing negative influence of learning the source distribution on target adaptation across various methods, resulting in a notable deterioration in model performance. }
\label{Fig_performance_dn_negative_ana}
\end{figure*}
\textbf{Experiment results.} We evaluated the time efficiency of SFDA and UDA using classical methods on single-source and multi-source transfer tasks. As shown in Fig. \ref{Fig_time_ana_intro} and \ref{Fig_performance_intro}, SFDA methods consistently outperform UDA in both efficiency and effectiveness.
SFDA achieves stable target domain performance within approximately 200 iterations, while UDA typically requires 1,000–5,000 iterations—5 to 25 times more. The absence of source data retraining in SFDA significantly reduces time requirements, enabling rapid transfer to downstream tasks. Additionally, SFDA consistently delivers superior performance across most transfer tasks, highlighting its adaptability and scalability.

\subsubsection{Learning objective analysis} 

The learning objectives of the UDA methods and SFDA methods are defined as follows.

\textbf{UDA methods:}
\begin{equation}  
\begin{aligned}  
\mathcal{L}_{UDA} = \mathop{\min}_{\theta} \mathbb{E}_{D_s} \mathcal{L}_{s}(\theta) + \lambda \mathbb{E}_{D_t} \mathcal{L}_{t}(\theta)\,,
\end{aligned} 
\label{Eq_uda_objectives}
\end{equation}
where $\theta$ is the parameters of the learning target model, $\mathcal{L}$ is the loss, and $\lambda$ is the relative weighting hyper-parameter between the source loss and target loss. Typically, the source loss is cross-entropy, while the target loss incorporates adversarial or minimal discrepancy terms.

\textbf{SFDA methods:}
\begin{equation}  
\begin{aligned}  
\mathcal{L}_{SFDA} = \mathop{\min}_{\theta} \mathbb{E}_{D_t} \mathcal{L}_{t}(\theta, M^s)\,,
\end{aligned} 
\label{Eq_sfda_objectives}
\end{equation}
where $M^s$ denotes the given source model, $\mathcal{L}_{SFDA}$ sometimes is self-trainning loss or constrative loss.

Eq. (\ref{Eq_uda_objectives}) shows that UDA prioritizes source data and treats target data as auxiliary, relying on distribution alignment to identify target samples when source and target domains are similar. However, significant domain differences can lead to learning interpolated distributions, degrading target performance. In contrast, Eq. (\ref{Eq_sfda_objectives}) highlights SFDA's focus on utilizing the source model's output to capture the target domain's structural characteristics. For small distribution gaps, this leads to better initial performance and guides learning effectively. For large gaps, avoiding repeated integration of dissimilar source data prevents negative transfer effects.

\textbf{Experiment results.} To evaluate the impact of UDA and SFDA objectives, we conducted the following experiments: (1) Base: Used classic MSFDA algorithms (e.g., SHOT$_{Avg}$, Decision, DATE) based on source models, focusing on target learning objectives (Eq. (\ref{Eq_sfda_objectives})). (2) Expanded Base: Building on experiment (1), we introduced visibility of data from a specific source domain, incorporating this visible source data into the target domain's learning objective using cross-entropy loss (Eq. (\ref{Eq_uda_objectives})). (3) Expanded Base with alignment strategies: Further incorporated UDA strategies (\emph{e.g.,} MMD loss, adversarial loss, confidence-based sample selection) to address domain differences. It is important to note that experiment (1) emphasizes learning within the target domain, while experiments (2) and (3) aim to learn an interpolated distribution between the source and target domains.

Fig. \ref{Fig_performance_shot_compare} demonstrates that when source-target distribution differences are minimal, target performance remains stable, with changes under one percentage point. However, Fig. \ref{Fig_performance_dn_compare} shows that large distribution gaps significantly harm target learning. For instance, integrating Quickdraw data can cause up to a 30\% performance drop in the target domain.
These findings underline the persistent challenge of mitigating adverse effects from distribution differences. Classical UDA alignment and sample selection techniques often fail to address the incompatibility of dissimilar source and target distributions, highlighting SFDA's robustness in such scenarios.

\begin{figure*}[h]
\begin{minipage}[b]{.999\linewidth}
\centering
\includegraphics[height=3.5cm,width=18cm]{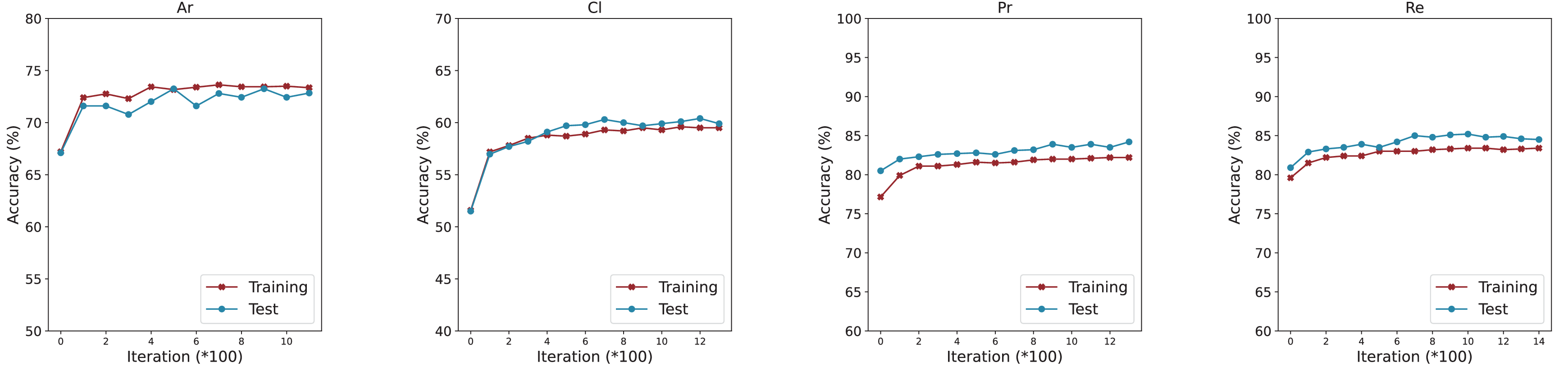}
\subcaption{Office-Home}
\end{minipage}
\begin{minipage}[b]{.999\linewidth}
\centering
\includegraphics[height=3.5cm,width=18cm]{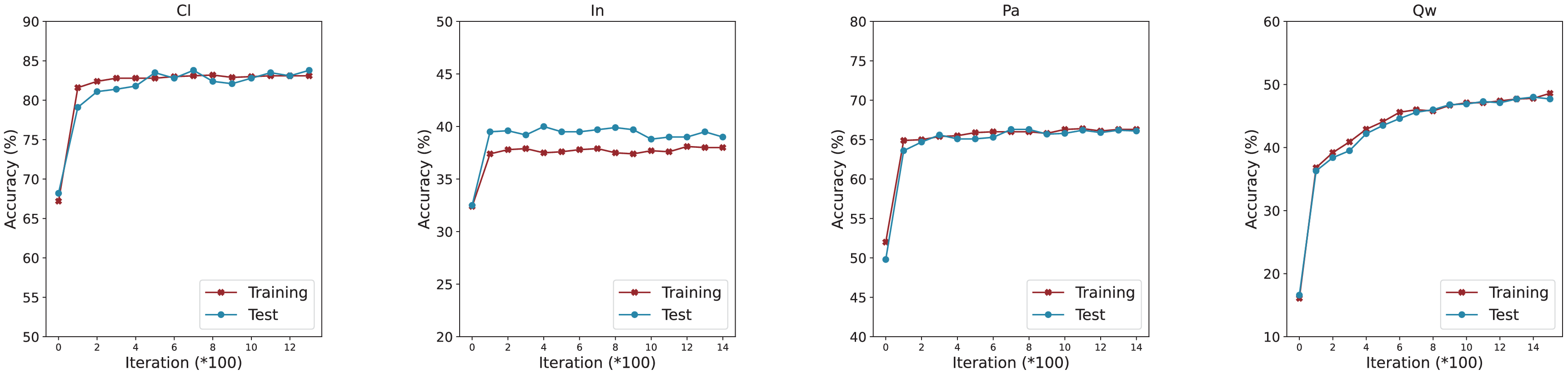}
\subcaption{DomainNet}
\end{minipage}
\caption{Analysis of overfitting of SHOT$_{Avg}$ (SFDA method) on the office-Home and DomainNet datasets.  The figure illustrates that training and test set accuracies are nearly identical, suggesting that source domain-independent adaptation algorithms based solely on the source domain model do not lead to overfitting. Models trained on the training set can generalize effectively to downstream target domain datasets. It is worth noting that the ratio of the training set to the test set is set at 9:1. }
\label{Fig_performance_overfitting}
\end{figure*}

\subsubsection{Negative transfer analysis}

In scenarios with significant source-target distribution disparities, the source model demonstrates greater resilience to negative transfer compared to source data. By abstracting key features without fully leveraging the source distribution, the model mitigates adverse effects associated with dissimilar distributions. SFDA methods utilize the source model’s outputs to dynamically adapt to the target domain, progressively aligning the model with the target distribution and minimizing the impact of source-target discrepancies. Furthermore, the source model focuses on domain-relevant features, reducing noise and computational overhead, thereby ensuring robust performance in challenging adaptation settings.

\textbf{Experiment results.}  To evaluate the impact of negative transfer, we compared traditional MSFDA methods (Decision and DATE) with and without visible source data on the DomainNet dataset. As illustrated in Fig. \ref{Fig_performance_dn_negative_ana}, MSFDA methods maintained stable performance in training. However, the integration of visible source data from dissimilar domains, such as Quickdraw, resulted in performance degradation of up to 30\% in the target domain. These findings emphasize the critical need to address negative transfer when dealing with significant distribution differences.

\subsubsection{Overfitting analysis}
A critical question in Source-Free Domain Adaptation (SFDA) is whether the high performance achieved through target adaptation alone results from overfitting to the target data. If true, models trained with SFDA methods may fail to generalize effectively to downstream tasks within the target domain. This section addresses this concern, providing both experimental evidence and analytical insights to demonstrate that SFDA methods do not exhibit overfitting.

\textbf{Experimental results:} To investigate potential overfitting in SFDA methods, we conducted experiments on datasets with varying distribution disparities. As shown in Fig. \ref{Fig_performance_overfitting}, MSFDA methods consistently maintain high accuracy across both training and test sets, indicating no overfitting. These results confirm the strong generalization capabilities of SFDA methods for downstream tasks in practical scenarios.

Two key factors contribute to this outcome. First, SFDA methods achieve rapid convergence, typically stabilizing within ~100 iterations (Fig. \ref{Fig_time_ana_intro}), which reduces the risk of excessive training and minimizes overfitting during adaptation. Second, SFDA approaches are designed to leverage the distributional knowledge encoded in the source model to uncover the latent structure of target data. Techniques like the clustering strategies in SHOT \cite{liangjian} and onBias \cite{shen2023balancing}, as well as the neighborhood structure exploration in NRC \cite{yang2021exploiting} and AaD \cite{yang2022attracting}, exemplify how SFDA methods aim to capture the intrinsic complexities of the target domain. By robustly modeling the overarching target structure, these approaches ensure stable performance without degradation.

\begin{figure}[t]
\centering
\includegraphics[height=3.5cm,width=7.5cm]{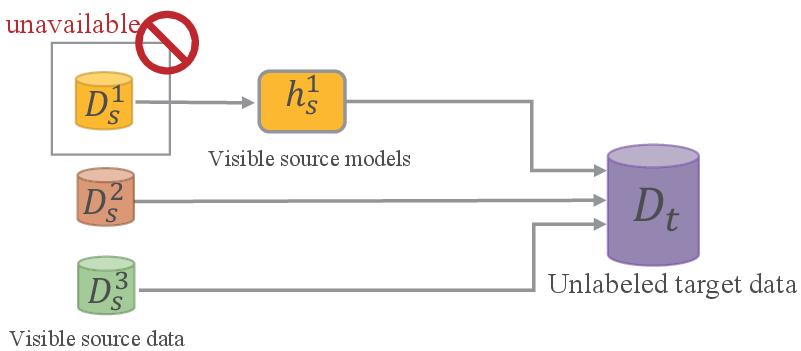}
\caption{The scenario of data-model fusion, where some source data are visible and some source models are visible. }
\label{data_model_fusion_intro}
\end{figure}

\section{Proposed Framework: Data-Model Fusion Scenario}

In this section, we introduce the novel data-model fusion scenario, which we propose for the first time as a practical framework to address the limitations of traditional UDA and SFDA paradigms in real-world applications. Existing methods often struggle to accommodate diverse collaboration settings where stakeholders share resources differently—some provide raw data, while others offer pre-trained models due to privacy constraints. Our framework bridges this gap by enabling seamless integration of data and models from multiple entities, allowing for privacy-preserving domain adaptation.

\subsection{Data-Model Fusion Scenario}

In this subsection, we extend the traditional problem settings of UDA and SFDA to define the data-model fusion scenario, a framework that reflects real-world resource-sharing complexities. Let $D_s^1 = \{x_i^{s_1}, y_i^{s_1}\}_{i=1}^{n_{s_1}}, \cdots, D_s^m = \{x_i^{s_m}, y_i^{s_m}\}_{i=1}^{n_{s_m}}$ denote the source domains, where $m$ represents the number of source domains and $n_{s_m}$ is the number of labeled samples in each domain. Similar to UDA and SFDA, $y \in \mathcal{Y} \in \mathcal{R}^K$ is the one-hot ground-truth label, where $K$ represents the total number of classes in the label set. Additionally, $D_t = \{x_i^t\}_{i=1}^{n_t}$ denotes the target domain, consisting of $n_t$ unlabeled samples that share the same label set $\mathcal{Y}$ as the source domains.

In the data-model fusion scenario, we introduce an additional complexity: not all source domains provide labeled data directly. Specifically, some source domains contribute labeled data, denoted as $D_s^i = \{x_i^{s_i}, y_i^{s_i}\}_{i=1}^{n_{s_i}}$, while others provide only pre-trained source models $M^j$, trained in a supervised manner on their respective source domains using cross-entropy loss. Each pre-trained source model $M^j$ comprises a feature extractor and a linear classifier.

Formally, the data-model fusion scenario includes both data-contributing source domains $\{D_s^i\}_{i=1}^{m_d}$ and model-contributing source domains $\{M^j\}_{j=1}^{m_m}$, where $m_d$ is the number of source domains contributing data, and $m_m$ is the number of source domains providing pre-trained models. The key objective of the data-model fusion scenario is to collaboratively leverage the available labeled data from $\{D_s^i\}_{i=1}^{m_d}$ and the pre-trained source models $\{M^j\}_{j=1}^{m_m}$ to effectively adapt to the unsupervised target domain $D_t$, all while adhering to privacy constraints and resource-sharing limitations.

This setting is motivated by real-world applications where resource-sharing constraints necessitate heterogeneous contributions. For example, in healthcare, public institutions may provide anonymized labeled datasets for research, while private hospitals constrained by privacy regulations contribute only pre-trained models. In this context, the data-model fusion scenario aims to optimize the use of these diverse resources to improve target domain adaptation performance. Similarly, in finance and public safety, combining anonymized transaction data with pre-trained fraud detection models can enhance collaborative performance while maintaining compliance with strict privacy protocols.
\begin{figure}[t]
\centering
\includegraphics[height=3.5cm,width=8cm]{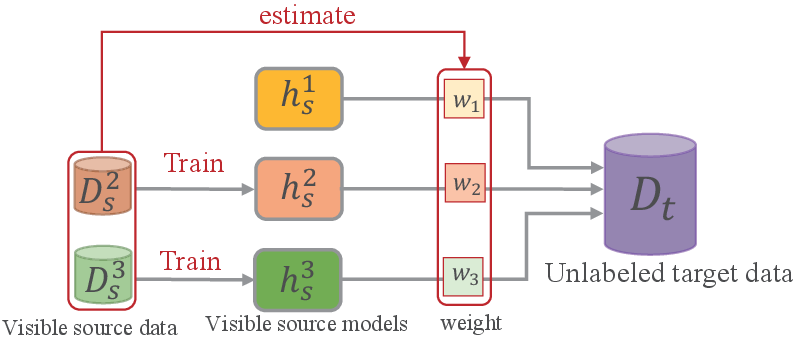}
\caption{Model Estimation and Adaptation Framework for data-model fusion scenario }
\label{data_model_fusion_framework}
\end{figure}

\subsection{Model Estimation and Adaptation Approach}

To effectively address the challenges of the data-model fusion scenario, we propose the Model Estimation and Adaptation (MEA) approach. Our strategy is grounded in the empirical observation that SFDA paradigms consistently outperform traditional UDA approaches in real-world settings. Consequently, MEA adopts a multi-source free domain adaptation strategy to handle heterogeneous resource-sharing environments. The core challenge in this framework lies in accurately estimating the contribution of each source model to the target domain. To tackle this, we propose a Proxy-Based Source Model Weight Estimation module, leveraging visible source domain data as proxies to assess and optimize source model weights. The MEA approach integrates this innovative module with a robust multi-source SFDA baseline to achieve superior performance in target domain adaptation.

\subsubsection{Adopting the SFDA Paradigm}
Our first step is to establish a solid multi-source free domain adaptation baseline, building on the proven superiority of SFDA methods over UDA. Each source model is pre-trained on its respective source domain using supervised learning. For instance, as shown in Fig.~\ref{data_model_fusion_framework}, model $h_s^2$ is trained on source data $D_s^2$, while model $h_s^3$ is trained on $D_s^3$. This can be formalized as:
\begin{equation}
H_s = \{H_s^i \, | \, 0 < i \leq m\},
\end{equation}
where $H_s$ represents the set of pre-trained source models, and $m$ denotes the number of source domains. If the source data $D_s^i$ is available, the model $h_s^i$ is trained on $D_s^i$; otherwise, we directly use the pre-trained model $h_s^i$. 

During adaptation, we integrate existing multi-source SFDA methods, such as SHOT$_{avg}$~\cite{liangjian} or DATE~\cite{han2023discriminability}, to minimize discrepancies between source models and the target domain. The loss function is defined as:
\begin{equation}  
\mathcal{L} (H;{\mathcal{D}_{t}}) = -{\mathbb{E}}_{x \in {\mathcal{D}}_{t}}{\sum}_{i=1}^m w_i(x_t) L(x_t),
\end{equation}
where $w_i(x_t)$ represents the weight of each source model $h_s^i$ for a target sample $x_t$, and $L(x_t)$ is the loss function (e.g., information maximization loss). This baseline establishes a foundation for adaptation by enabling collaboration across multiple source models without requiring access to source data, effectively reducing negative transfer.

\subsubsection{Proxy-Based Source Model Weight Estimation}
The critical challenge in the data-model fusion scenario is to accurately estimate the contribution of each source model to the target domain. Traditional SFDA methods assume equal weights for source models or rely solely on target domain data, which may be insufficient in complex, resource-sharing environments. To overcome this limitation, we propose a Proxy-Based Source Model Weight Estimation strategy, leveraging visible source domain data as proxy domains to evaluate the performance of each source model. 

\textit{Step 1: Accuracy-Based Weight Estimation.}
In the absence of labeled target data, we evaluate the accuracy of each source model on other visible source domains as a proxy. High accuracy on proxy domains indicates a source model's strong ability to extract domain-invariant features, which enhances its relevance to the target domain. Specifically, the weight of a source model $h_s^i$ is calculated as:
\begin{equation}  
w_i^s = \frac{A(w_i)}{{\sum}_{i=1}^m A(w_i)},
\end{equation}
where $A(w_i)$ represents the accuracy of model $h_s^i$ on alternative visible source domains. For instance, $A(w_1)$ reflects the average accuracy of model $h_s^1$ on datasets $D_s^2$ and $D_s^3$, while $A(w_2)$ is calculated based on $D_s^1$ and $D_s^3$. Importantly, we exclude a model's own training data (\emph{e.g.,} $D_s^2$ for $h_s^2$) to mitigate bias and ensure unbiased weight estimation.

This proxy-based approach addresses scenarios where partial source data are visible, enabling an effective estimation of model relevance without requiring direct target supervision. However, in cases with only one source domain, this strategy relies on multi-source SFDA techniques for adaptation.

\textit{Step 2: Confidence-Based Weight Estimation.}
In addition to proxy domain accuracy, we incorporate target domain information by analyzing the confidence of each source model on target domain samples. Specifically, we compute the average confidence score for each source model, defined as the highest predicted probability across all target samples:
\begin{equation}
w_i^t = \frac{C(w_i)}{{\sum}_{i=1}^m C(w_i)},
\end{equation}
where $C(w_i) = \frac{{\sum}_{j=1}^{n_t} \delta(h_s^i(x_t^j))}{n_t}$ represents the average softmax confidence scores of model $h_s^i$ across $n_t$ target samples. Higher confidence levels indicate a stronger alignment of the source model with the target domain.

\textit{Step 3: Combined Weight Estimation.}
The final weight of each source model is calculated as a combination of proxy domain accuracy and target domain confidence:
\begin{equation}  
w_i = w_i^t + \lambda w_i^s,
\end{equation}
where $\lambda$ is a hyperparameter balancing the contributions of the two components. By incorporating both proxy domain and target domain information, this strategy provides a robust and adaptive estimation of source model relevance in the data-model fusion scenario.

\subsubsection{Integration and Practical Implications}
The Proxy-Based Source Model Weight Estimation module is seamlessly integrated into the MEA framework, enabling efficient adaptation to the target domain while respecting resource-sharing constraints. By leveraging visible source data as proxies, the MEA approach addresses key challenges in the data-model fusion scenario, ensuring robust adaptation performance in diverse, privacy-sensitive environments. Our empirical results validate the effectiveness of this strategy, demonstrating consistent improvements over existing multi-SFDA and UDA methods, thereby establishing a new benchmark for domain adaptation in real-world applications.

\begin{table}
\centering
	\caption{Accuracy (\%) on DomainNet (ResNet50) in existing classical multi-UDA methods, multi-SFDA methods and our MEA methods in data-model fusion scenario (‘SF’ in tables denotes source data free, \emph{i.e.,} adaptation without source data).}
	\label{tab:performance_comparison}
    \scalebox{.85}{
    \renewcommand{\arraystretch}{1.2}
    \begin{tabular}{l|c|c|ccccc}
    \toprule
    Categories&Method&SF&C&I&P&Q&Avg\cr
    \midrule
    None& Source-only&$\checkmark$&67.2&32.4&51.8&16.2 & 41.9\cr
    \midrule
    \multirow{4}{*}{Multi-UDA} 
        & MFSAN \cite{zhu2019aligning} & $\times$ & 77.2 & 36.7 & 63.1 & 41.3 & 54.6 \\
       & M$^3$SDA \cite{peng2019moment} & $\times$ & 59.7 & 29.4 & 48.8 & 16.0 & 38.5 \\ 
       & MIAN  \cite{park2021information} &$\times$ & 76.5&37.9&60.6&36.2&52.8 \\
       & MSCLDA \cite{li2021multi} &$\times$ &77.3& 36.5&63.9&40.2&54.5\\
    \midrule
    \multirow{4}{*}{Multi-SFDA} 
       
        & SHOT$_{avg}$\cite{liangjian} & $\checkmark$ & 83.7 & 38.4 &  65.9 & 46.6 & 58.7\\
       & MEA + SHOT$_{avg}$ & partial & \textbf{84.2} & 39.6 &  66.3 & \textbf{47.2} & 59.3\\
       \cmidrule{2-8}
       & DATE \cite{han2023discriminability} &$\checkmark$ & 82.6 & 39.7 &  67.4 & 44.4 & 58.5\\
       & MEA + DATE & partial & 82.8 & \textbf{40.4}  & \textbf{67.9} & 46.3 & \textbf{59.4}\\
   \toprule
	\end{tabular}}
	\label{Result_MEA}
\end{table}
\subsubsection{Experimental analysis on MEA framework}
To evaluate the MEA framework, we conducted experiments on the challenging DomainNet dataset, one of the largest benchmarks for domain adaptation, using two state-of-the-art multi-SFDA methods, SHOT$_{avg}$ and DATE, as baselines. We replaced their standard weight estimation strategies with our proposed proxy-based source model weight estimation approach, leveraging data from two randomly selected domains as visible sources.

Results in Table~\ref{Result_MEA} demonstrate that the MEA framework consistently enhances the performance of both SHOT${avg}$ and DATE across all target domains. Specifically, MEA improves SHOT${avg}$ by 0.5\% on average (from 58.7\% to 59.3\%) and DATE by 0.9\% on average (from 58.5\% to 59.4\%). These improvements are consistent across all individual domains, highlighting the robustness of MEA in diverse target adaptation scenarios. Moreover, MEA outperforms SOTA multi-UDA methods, such as MFSAN and MIAN, by a large margin (e.g., achieving an average accuracy of 59.4\% compared to 54.6\% and 52.8\% for MFSAN and MIAN, respectively). 

Importantly, these results validate the generality of the MEA framework. By seamlessly integrating with different multi-SFDA methods, MEA demonstrates its adaptability and effectiveness in various domain adaptation pipelines. For example, on domain Q, MEA consistently outperforms the baselines, with improvements in both SHOT$_{avg}$ (from 46.6\% to 47.2\%) and DATE (from 44.4\% to 46.3\%). These consistent improvements underscore MEA’s ability to enhance both existing multi-SFDA methods and the overall adaptation process.

\section{Conclusion}

This paper comprehensively evaluated Unsupervised Domain Adaptation (UDA) and Source-Free Domain Adaptation (SFDA) in addressing domain shifts, drawing on predictive coding theory and extensive experiments on benchmark datasets. Our findings demonstrate that SFDA consistently outperforms UDA, particularly in scenarios with significant source-target distribution gaps. This is attributed to SFDA’s ability to mitigate negative transfer, resist overfitting, and operate without source data, making it especially suitable for privacy-sensitive and resource-constrained applications. In addition, we introduced the novel data-model fusion scenario, a practical framework that extends the adaptability of SFDA to complex, real-world settings where entities share resources heterogeneously—some contributing raw data while others provide pre-trained models. To address the challenges of this scenario, we proposed the Model Estimation and Adaptation (MEA) framework, which incorporates a proxy-based source model weight estimation strategy. By leveraging visible source data as proxies, the MEA framework effectively evaluates and integrates source model contributions, achieving superior performance in complex adaptation tasks.

\bibliographystyle{ieeetr}
\bibliography{main}

\begin{thebibliography}{10}

\bibitem{ferenc2020deep}
R.~Ferenc, D.~B{\'a}n, T.~Gr{\'o}sz, and T.~Gyim{\'o}thy, ``Deep learning in static, metric-based bug prediction,'' {\em Array}, vol.~6, p.~100021, 2020.

\bibitem{zhang2015digital}
Y.~Zhang, P.~Li, Y.~Jin, and Y.~Choe, ``A digital liquid state machine with biologically inspired learning and its application to speech recognition,'' {\em IEEE transactions on neural networks and learning systems}, vol.~26, no.~11, pp.~2635--2649, 2015.

\bibitem{li2015twitter}
H.~Li, N.~Guevara, N.~Herndon, D.~Caragea, K.~Neppalli, C.~Caragea, A.~C. Squicciarini, and A.~H. Tapia, ``Twitter mining for disaster response: A domain adaptation approach.,'' in {\em ISCRAM}, 2015.

\bibitem{long2017conditional}
M.~Long, Z.~Cao, J.~Wang, and M.~I. Jordan, ``Conditional adversarial domain adaptation,'' {\em arXiv preprint arXiv:1705.10667}, 2017.

\bibitem{long2015learning}
M.~Long, Y.~Cao, J.~Wang, and M.~Jordan, ``Learning transferable features with deep adaptation networks,'' in {\em International conference on machine learning}, pp.~97--105, PMLR, 2015.

\bibitem{ben2007analysis}
S.~Ben-David, J.~Blitzer, K.~Crammer, F.~Pereira, {\em et~al.}, ``Analysis of representations for domain adaptation,'' {\em Advances in neural information processing systems}, vol.~19, p.~137, 2007.

\bibitem{ben2010theory}
S.~Ben-David, J.~Blitzer, K.~Crammer, A.~Kulesza, F.~Pereira, and J.~W. Vaughan, ``A theory of learning from different domains,'' {\em Machine learning}, vol.~79, pp.~151--175, 2010.

\bibitem{chen2020homm}
C.~Chen, Z.~Fu, Z.~Chen, S.~Jin, Z.~Cheng, X.~Jin, and X.-S. Hua, ``Homm: Higher-order moment matching for unsupervised domain adaptation,'' in {\em Proceedings of the AAAI conference on artificial intelligence}, vol.~34, pp.~3422--3429, 2020.

\bibitem{peng2019moment}
X.~Peng, Q.~Bai, X.~Xia, Z.~Huang, K.~Saenko, and B.~Wang, ``Moment matching for multi-source domain adaptation,'' in {\em Proceedings of the IEEE/CVF International Conference on Computer Vision}, pp.~1406--1415, 2019.

\bibitem{zellinger2019robust}
W.~Zellinger, B.~A. Moser, T.~Grubinger, E.~Lughofer, T.~Natschl{\"a}ger, and S.~Saminger-Platz, ``Robust unsupervised domain adaptation for neural networks via moment alignment,'' {\em Information Sciences}, vol.~483, pp.~174--191, 2019.

\bibitem{nguyen2024class}
T.~Nguyen, V.~Nguyen, T.~Le, H.~Zhao, Q.~H. Tran, and D.~Phung, ``A class-aware optimal transport approach with higher-order moment matching for unsupervised domain adaptation,'' {\em arXiv preprint arXiv:2401.15952}, 2024.

\bibitem{qiang2021robust}
W.~Qiang, J.~Li, C.~Zheng, B.~Su, and H.~Xiong, ``Robust local preserving and global aligning network for adversarial domain adaptation,'' {\em IEEE Transactions on Knowledge and Data Engineering}, vol.~35, no.~3, pp.~3014--3029, 2021.

\bibitem{saito2017adversarial}
K.~Saito, Y.~Ushiku, T.~Harada, and K.~Saenko, ``Adversarial dropout regularization,'' {\em arXiv preprint arXiv:1711.01575}, 2017.

\bibitem{tzeng2017adversarial}
E.~Tzeng, J.~Hoffman, K.~Saenko, and T.~Darrell, ``Adversarial discriminative domain adaptation,'' in {\em Proceedings of the IEEE conference on computer vision and pattern recognition}, pp.~7167--7176, 2017.

\bibitem{Tang_2020_CVPR}
H.~Tang, K.~Chen, and K.~Jia, ``Unsupervised domain adaptation via structurally regularized deep clustering,'' in {\em Proceedings of the IEEE/CVF Conference on Computer Vision and Pattern Recognition (CVPR)}, June 2020.

\bibitem{liangjian}
J.~Liang, D.~Hu, and J.~Feng, ``Do we really need to access the source data ? source hypothesis transfer for unsupervised domain adaptation,'' in {\em International Conference on Machine Learning}, pp.~6028--6039, PMLR, 2020.

\bibitem{liang2021source}
J.~Liang, D.~Hu, Y.~Wang, R.~He, and J.~Feng, ``Source data-absent unsupervised domain adaptation through hypothesis transfer and labeling transfer,'' {\em IEEE Transactions on Pattern Analysis and Machine Intelligence}, 2021.

\bibitem{kundu2022balancing}
J.~N. Kundu, A.~R. Kulkarni, S.~Bhambri, D.~Mehta, S.~A. Kulkarni, V.~Jampani, and V.~B. Radhakrishnan, ``Balancing discriminability and transferability for source-free domain adaptation,'' in {\em International Conference on Machine Learning}, pp.~11710--11728, PMLR, 2022.

\bibitem{DBLP:journals/jcst/TianMZPX21}
Q.~Tian, C.~Ma, F.~Zhang, S.~Peng, and H.~Xue, ``Source-free unsupervised domain adaptation with sample transport learning,'' {\em J. Comput. Sci. Technol.}, vol.~36, no.~3, pp.~606--616, 2021.

\bibitem{yang2021exploiting}
S.~Yang, J.~van~de Weijer, L.~Herranz, S.~Jui, {\em et~al.}, ``Exploiting the intrinsic neighborhood structure for source-free domain adaptation,'' {\em Advances in Neural Information Processing Systems}, vol.~34, 2021.

\bibitem{yang2021generalized}
S.~Yang, Y.~Wang, J.~van~de Weijer, L.~Herranz, and S.~Jui, ``Generalized source-free domain adaptation,'' in {\em Proceedings of the IEEE/CVF International Conference on Computer Vision}, pp.~8978--8987, 2021.

\bibitem{yang2022attracting}
S.~Yang, S.~Jui, J.~van~de Weijer, {\em et~al.}, ``Attracting and dispersing: A simple approach for source-free domain adaptation,'' {\em Advances in Neural Information Processing Systems}, vol.~35, pp.~5802--5815, 2022.

\bibitem{yang2022one}
S.~Yang, Y.~Wang, K.~Wang, S.~Jui, and J.~van~de Weijer, ``One ring to bring them all: Towards open-set recognition under domain shift,'' {\em arXiv preprint arXiv: 2206.03600}, 2022.

\bibitem{rao1999predictive}
R.~P. Rao and D.~H. Ballard, ``Predictive coding in the visual cortex: a functional interpretation of some extra-classical receptive-field effects,'' {\em Nature neuroscience}, vol.~2, no.~1, pp.~79--87, 1999.

\bibitem{spratling2017review}
M.~W. Spratling, ``A review of predictive coding algorithms,'' {\em Brain and cognition}, vol.~112, pp.~92--97, 2017.

\bibitem{long2017deep}
M.~Long, H.~Zhu, J.~Wang, and M.~I. Jordan, ``Deep transfer learning with joint adaptation networks,'' in {\em International conference on machine learning}, pp.~2208--2217, PMLR, 2017.

\bibitem{zhang2019bridging}
Y.~Zhang, T.~Liu, M.~Long, and M.~Jordan, ``Bridging theory and algorithm for domain adaptation,'' in {\em International Conference on Machine Learning}, pp.~7404--7413, PMLR, 2019.

\bibitem{xu2019larger}
R.~Xu, G.~Li, J.~Yang, and L.~Lin, ``Larger norm more transferable: An adaptive feature norm approach for unsupervised domain adaptation,'' in {\em Proceedings of the IEEE/CVF International Conference on Computer Vision}, pp.~1426--1435, 2019.

\bibitem{ganin2015unsupervised}
Y.~Ganin and V.~Lempitsky, ``Unsupervised domain adaptation by backpropagation,'' in {\em International conference on machine learning}, pp.~1180--1189, PMLR, 2015.

\bibitem{xu2018deep}
R.~Xu, Z.~Chen, W.~Zuo, J.~Yan, and L.~Lin, ``Deep cocktail network: Multi-source unsupervised domain adaptation with category shift,'' in {\em Proceedings of the IEEE conference on computer vision and pattern recognition}, pp.~3964--3973, 2018.

\bibitem{zhu2019aligning}
Y.~Zhu, F.~Zhuang, and D.~Wang, ``Aligning domain-specific distribution and classifier for cross-domain classification from multiple sources,'' in {\em Proceedings of the AAAI conference on artificial intelligence}, vol.~33, pp.~5989--5996, 2019.

\bibitem{nguyen2021stem}
V.-A. Nguyen, T.~Nguyen, T.~Le, Q.~H. Tran, and D.~Phung, ``Stem: An approach to multi-source domain adaptation with guarantees,'' in {\em Proceedings of the IEEE/CVF International Conference on Computer Vision}, pp.~9352--9363, 2021.

\bibitem{yang2020curriculum}
L.~Yang, Y.~Balaji, S.-N. Lim, and A.~Shrivastava, ``Curriculum manager for source selection in multi-source domain adaptation,'' in {\em Computer Vision--ECCV 2020: 16th European Conference, Glasgow, UK, August 23--28, 2020, Proceedings, Part XIV 16}, pp.~608--624, Springer, 2020.

\bibitem{park2021information}
G.~Y. Park and S.~W. Lee, ``Information-theoretic regularization for multi-source domain adaptation,'' in {\em Proceedings of the IEEE/CVF International Conference on Computer Vision}, pp.~9214--9223, 2021.

\bibitem{venkat2020your}
N.~Venkat, J.~N. Kundu, D.~Singh, A.~Revanur, {\em et~al.}, ``Your classifier can secretly suffice multi-source domain adaptation,'' {\em Advances in Neural Information Processing Systems}, vol.~33, pp.~4647--4659, 2020.

\bibitem{zhao2020multi}
S.~Zhao, G.~Wang, S.~Zhang, Y.~Gu, Y.~Li, Z.~Song, P.~Xu, R.~Hu, H.~Chai, and K.~Keutzer, ``Multi-source distilling domain adaptation,'' in {\em Proceedings of the AAAI Conference on Artificial Intelligence}, vol.~34, pp.~12975--12983, 2020.

\bibitem{li2021multi}
K.~Li, J.~Lu, H.~Zuo, and G.~Zhang, ``Multi-source contribution learning for domain adaptation,'' {\em IEEE Transactions on Neural Networks and Learning Systems}, vol.~33, no.~10, pp.~5293--5307, 2021.

\bibitem{wen2020domain}
J.~Wen, R.~Greiner, and D.~Schuurmans, ``Domain aggregation networks for multi-source domain adaptation,'' in {\em International conference on machine learning}, pp.~10214--10224, PMLR, 2020.

\bibitem{wang2019tmda}
H.~Wang, W.~Yang, Z.~Lin, and Y.~Yu, ``Tmda: Task-specific multi-source domain adaptation via clustering embedded adversarial training,'' in {\em 2019 IEEE International Conference on Data Mining (ICDM)}, pp.~1372--1377, IEEE, 2019.

\bibitem{huang2021model}
J.~Huang, D.~Guan, A.~Xiao, and S.~Lu, ``Model adaptation: Historical contrastive learning for unsupervised domain adaptation without source data,'' {\em Advances in Neural Information Processing Systems}, vol.~34, 2021.

\bibitem{xia2021adaptive}
H.~Xia, H.~Zhao, and Z.~Ding, ``Adaptive adversarial network for source-free domain adaptation,'' in {\em Proceedings of the IEEE/CVF International Conference on Computer Vision}, pp.~9010--9019, 2021.

\bibitem{li2020model}
R.~Li, Q.~Jiao, W.~Cao, H.-S. Wong, and S.~Wu, ``Model adaptation: Unsupervised domain adaptation without source data,'' in {\em Proceedings of the IEEE/CVF Conference on Computer Vision and Pattern Recognition}, pp.~9641--9650, 2020.

\bibitem{kurmi2021domain}
V.~K. Kurmi, V.~K. Subramanian, and V.~P. Namboodiri, ``Domain impression: A source data free domain adaptation method,'' in {\em Proceedings of the IEEE/CVF Winter Conference on Applications of Computer Vision}, pp.~615--625, 2021.

\bibitem{guan2021domain}
D.~Guan, J.~Huang, A.~Xiao, and S.~Lu, ``Domain adaptive video segmentation via temporal consistency regularization,'' in {\em Proceedings of the IEEE/CVF International Conference on Computer Vision}, pp.~8053--8064, 2021.

\bibitem{ye2022alleviating}
Y.~Ye, Z.~Liu, Y.~Zhang, J.~Li, and H.~Shen, ``Alleviating style sensitivity then adapting: Source-free domain adaptation for medical image segmentation,'' in {\em Proceedings of the 30th ACM International Conference on Multimedia}, pp.~1935--1944, 2022.

\bibitem{yang2022source}
C.~Yang, X.~Guo, Z.~Chen, and Y.~Yuan, ``Source free domain adaptation for medical image segmentation with fourier style mining,'' {\em Medical Image Analysis}, vol.~79, p.~102457, 2022.

\bibitem{ahmed2021unsupervised}
S.~M. Ahmed, D.~S. Raychaudhuri, S.~Paul, S.~Oymak, and A.~K. Roy-Chowdhury, ``Unsupervised multi-source domain adaptation without access to source data,'' in {\em Proceedings of the IEEE/CVF conference on computer vision and pattern recognition}, pp.~10103--10112, 2021.

\bibitem{dong2021confident}
J.~Dong, Z.~Fang, A.~Liu, G.~Sun, and T.~Liu, ``Confident anchor-induced multi-source free domain adaptation,'' {\em Advances in Neural Information Processing Systems}, vol.~34, pp.~2848--2860, 2021.

\bibitem{pei2024evidential}
J.~Pei, A.~Men, Y.~Liu, X.~Zhuang, and Q.~Chen, ``Evidential multi-source-free unsupervised domain adaptation,'' {\em IEEE Transactions on Pattern Analysis and Machine Intelligence}, 2024.

\bibitem{shen2023balancing}
M.~Shen, Y.~Bu, and G.~W. Wornell, ``On balancing bias and variance in unsupervised multi-source-free domain adaptation,'' in {\em International Conference on Machine Learning}, pp.~30976--30991, PMLR, 2023.

\bibitem{li2023target}
G.~Li, Q.~Zhang, P.~Wang, R.~He, and C.~Wu, ``Target-discriminability-induced multi-source-free domain adaptation,'' in {\em 2023 IEEE International Conference on Image Processing (ICIP)}, pp.~76--80, IEEE, 2023.

\bibitem{han2023discriminability}
Z.~Han, Z.~Zhang, F.~Wang, R.~He, W.~Su, X.~Xi, and Y.~Yin, ``Discriminability and transferability estimation: a bayesian source importance estimation approach for multi-source-free domain adaptation,'' in {\em Proceedings of the AAAI Conference on Artificial Intelligence}, vol.~37, pp.~7811--7820, 2023.

\bibitem{saenko2010adapting}
K.~Saenko, B.~Kulis, M.~Fritz, and T.~Darrell, ``Adapting visual category models to new domains,'' in {\em European conference on computer vision}, pp.~213--226, Springer, 2010.

\bibitem{venkateswara2017deep}
H.~Venkateswara, J.~Eusebio, S.~Chakraborty, and S.~Panchanathan, ``Deep hashing network for unsupervised domain adaptation,'' in {\em Proceedings of the IEEE conference on computer vision and pattern recognition}, pp.~5018--5027, 2017.

\bibitem{fang2013unbiased}
C.~Fang, Y.~Xu, and D.~N. Rockmore, ``Unbiased metric learning: On the utilization of multiple datasets and web images for softening bias,'' in {\em Proceedings of the IEEE International Conference on Computer Vision}, pp.~1657--1664, 2013.

\bibitem{beery2018recognition}
S.~Beery, G.~Van~Horn, and P.~Perona, ``Recognition in terra incognita,'' in {\em Proceedings of the European conference on computer vision (ECCV)}, pp.~456--473, 2018.

\bibitem{gulrajani2020search}
I.~Gulrajani and D.~Lopez-Paz, ``In search of lost domain generalization,'' {\em arXiv preprint arXiv:2007.01434}, 2020.

\bibitem{he2016deep}
K.~He, X.~Zhang, S.~Ren, and J.~Sun, ``Deep residual learning for image recognition,'' in {\em Proceedings of the IEEE conference on computer vision and pattern recognition}, pp.~770--778, 2016.

\bibitem{ganin2016domain}
Y.~Ganin, E.~Ustinova, H.~Ajakan, P.~Germain, H.~Larochelle, F.~Laviolette, M.~Marchand, and V.~Lempitsky, ``Domain-adversarial training of neural networks,'' {\em The journal of machine learning research}, vol.~17, no.~1, pp.~2096--2030, 2016.

\bibitem{saito2018maximum}
K.~Saito, K.~Watanabe, Y.~Ushiku, and T.~Harada, ``Maximum classifier discrepancy for unsupervised domain adaptation,'' in {\em Proceedings of the IEEE conference on computer vision and pattern recognition}, pp.~3723--3732, 2018.

\bibitem{jin2020minimum}
Y.~Jin, X.~Wang, M.~Long, and J.~Wang, ``Minimum class confusion for versatile domain adaptation,'' in {\em Computer Vision--ECCV 2020: 16th European Conference, Glasgow, UK, August 23--28, 2020, Proceedings, Part XXI 16}, pp.~464--480, Springer, 2020.

\bibitem{pei2023uncertainty}
J.~Pei, Z.~Jiang, A.~Men, L.~Chen, Y.~Liu, and Q.~Chen, ``Uncertainty-induced transferability representation for source-free unsupervised domain adaptation,'' {\em IEEE Transactions on Image Processing}, vol.~32, pp.~2033--2048, 2023.

\bibitem{qiu2021source}
Z.~Qiu, Y.~Zhang, H.~Lin, S.~Niu, Y.~Liu, Q.~Du, and M.~Tan, ``Source-free domain adaptation via avatar prototype generation and adaptation,'' {\em arXiv preprint arXiv:2106.15326}, 2021.

\bibitem{wang2020learning}
H.~Wang, M.~Xu, B.~Ni, and W.~Zhang, ``Learning to combine: Knowledge aggregation for multi-source domain adaptation,'' in {\em Computer Vision--ECCV 2020: 16th European Conference, Glasgow, UK, August 23--28, 2020, Proceedings, Part VIII 16}, pp.~727--744, Springer, 2020.

\bibitem{li2021t}
R.~Li, X.~Jia, J.~He, S.~Chen, and Q.~Hu, ``T-svdnet: Exploring high-order prototypical correlations for multi-source domain adaptation,'' in {\em Proceedings of the IEEE/CVF International Conference on Computer Vision}, pp.~9991--10000, 2021.

\bibitem{nguyen2021most}
T.~Nguyen, T.~Le, H.~Zhao, Q.~H. Tran, T.~Nguyen, and D.~Phung, ``Most: Multi-source domain adaptation via optimal transport for student-teacher learning,'' in {\em Uncertainty in Artificial Intelligence}, pp.~225--235, PMLR, 2021.

\end{thebibliography}

\end{document}